\documentclass[conference]{IEEEtran}
\IEEEoverridecommandlockouts
\usepackage{cite}
\usepackage{amsmath,amssymb,amsfonts}
\usepackage{algorithmic}
\usepackage{graphicx}
\usepackage{amsmath}
\usepackage{booktabs}
\usepackage{algorithm}
\usepackage{algorithmic}
\usepackage{amsfonts}
\usepackage{multirow}
\usepackage{makecell}
\usepackage{subfigure}
\usepackage{textcomp}
\usepackage{xcolor}
\usepackage{url}
\usepackage{bm}
\usepackage{bbm}

\usepackage{fancyhdr}
\usepackage{algorithm}
\usepackage{listings}
\usepackage{etoolbox}
\makeatletter
\AfterEndEnvironment{algorithm}{\let\@algcomment\relax}
\AtEndEnvironment{algorithm}{\kern2pt\hrule\relax\vskip3pt\@algcomment}
\let\@algcomment\relax
\newcommand\algcomment[1]{\def\@algcomment{\footnotesize#1}}
\renewcommand\fs@ruled{\def\@fs@cfont{\bfseries}\let\@fs@capt\floatc@ruled
 \def\@fs@pre{\hrule height.8pt depth0pt \kern2pt}%
 \def\@fs@post{}%
 \def\@fs@mid{\kern2pt\hrule\kern2pt}%
 \let\@fs@iftopcapt\iftrue}
\makeatother

\newcommand{\eg}{\textit{e}.\textit{g}.}
\def\BibTeX{{\rm B\kern-.05em{\sc i\kern-.025em b}\kern-.08em
    T\kern-.1667em\lower.7ex\hbox{E}\kern-.125emX}}

\DeclareRobustCommand*{\IEEEauthorrefmark}[1]{%
    \raisebox{0pt}[0pt][0pt]{\textsuperscript{\footnotesize\ensuremath{#1}}}}

\IEEEoverridecommandlockouts

\begin{document}

\title{ID-MixGCL: Identity Mixup for Graph Contrastive Learning
}

\author{
	\IEEEauthorblockN{
		Gehang Zhang\IEEEauthorrefmark{1}\IEEEauthorrefmark{2}, 
		Bowen Yu\IEEEauthorrefmark{3}, 
		Jiangxia Cao\IEEEauthorrefmark{4}, 
		Xinghua Zhang\IEEEauthorrefmark{1}\IEEEauthorrefmark{2},
            Jiawei Sheng\IEEEauthorrefmark{1}$^*$\thanks{$^*$ Jiawei Sheng is the corresponding author.},
            Chuan Zhou\IEEEauthorrefmark{5},
            and Tingwen Liu\IEEEauthorrefmark{1}\IEEEauthorrefmark{2}
         } 
	\IEEEauthorblockA{\IEEEauthorrefmark{1}Institute of Information Engineering, Chinese Academy of Sciences, Beijing, China}
	\IEEEauthorblockA{\IEEEauthorrefmark{2}School of Cyber Security, University of Chinese Academy of Sciences, Beijing, China}
	\IEEEauthorblockA{\IEEEauthorrefmark{3} DAMO Academy, Alibaba Group \IEEEauthorrefmark{4} Kuaishou, Beijing, China} 
	\IEEEauthorblockA{\IEEEauthorrefmark{5}Academy of Mathematics and Systems Science, Chinese Academy of Sciences, Beijing, China \\ \{zhanggehang, zhangxinghua, shengjiawei,  liutingwen\}@iie.ac.cn, \{yubowen.ph, caojiangxia\}@gmail.com,\ zhouchuan@amss.ac.cn}
}

\maketitle
\thispagestyle{fancy}


\cfoot{}

\renewcommand{\headrulewidth}{0mm}

\begin{abstract}
Graph contrastive learning (GCL) has recently achieved substantial advancements. 
Existing GCL approaches compare two different ``views'' of the same graph in order to learn node/graph representations.
The underlying assumption of these studies is that the graph augmentation strategy is capable of generating several different graph views such that the graph views are structurally different but semantically similar to the original graphs, and thus the ground-truth labels of the original and augmented graph/nodes can be regarded identical in contrastive learning. 
However, we observe that this assumption does not always hold. For instance, the deletion of a super-node within a social network can exert a substantial influence on the partitioning of communities for other nodes. Similarly, any perturbation to nodes or edges in a molecular graph will change the labels of the graph.
Therefore, we believe that augmenting the graph,  accompanied by an adaptation of the labels used for the contrastive loss, will facilitate the encoder to learn a better representation. 
Based on this idea, we propose ID-MixGCL, which allows the simultaneous interpolation of input nodes and corresponding identity labels to obtain soft-confidence samples, with a controllable degree of change, leading to the capture of fine-grained representations from self-supervised training on unlabeled graphs.
Experimental results demonstrate that ID-MixGCL improves performance on graph classification and node classification tasks, as demonstrated by significant improvements on the Cora, IMDB-B, IMDB-M, and PROTEINS datasets compared to state-of-the-art techniques, by 3-29\% absolute points. 
\end{abstract}

\begin{IEEEkeywords}
Graph Representation Learning; Contrastive Learning; Self-supervised learning;
\end{IEEEkeywords}

\section{Introduction}
\begin{figure}[t!]
       \begin{center}
       \includegraphics[width=0.46\textwidth]{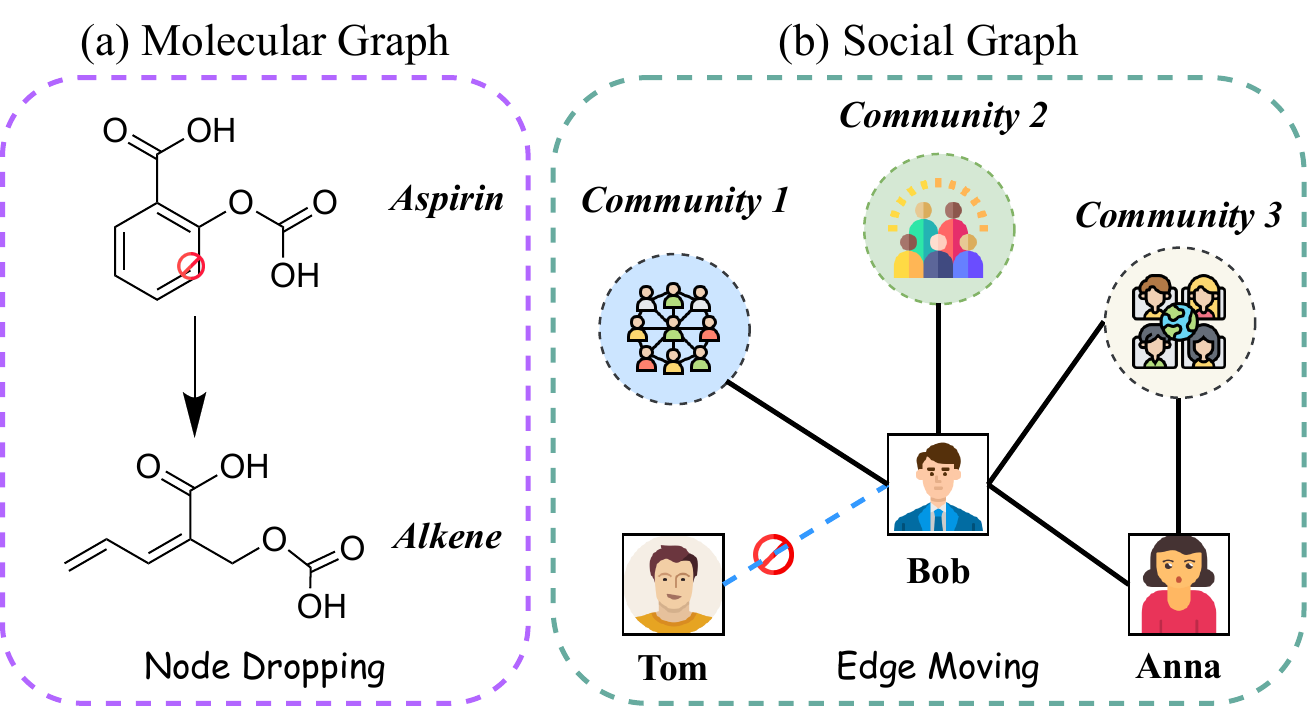}
       \caption{An example to illustrate that augmentation on the graph can unexpectedly change the type of graph or nodes. The left panel shows that removing a carbon atom from the phenyl ring of aspirin causes the molecule to become an alkene chain, and the right panel shows that removing the link between Tom and Bob causes Tom to become an isolated node. This motivates us to generate `soft-confidence' contrastive samples.
       }
        \label{motivation-visual}
	\end{center}
\end{figure}

Graph data is ubiquitous in human society, from the micro-level chemical molecule, biological protein to the macro-level traffic network, advertising clicks, thereby how to analyze those graph data better is attracted a surge of interest in past decades.
In recent years, the research attention concentrate on the Graph Neural Networks (GNNs), which aims to optimize the graph encoder to transform the node attributes and graph structure into a dense and low-dimensional feature space.
As a promising way to train such a graph encoder, self-supervised contrastive learning achieves impressive progress, where the key idea is to first generate several augmented graph views and then utilize the positive and negative node/graph pairs as supervised signals.

By extending the above graph contrastive learning (GCL) framework, a number of studies have been proposed, such as GraphCL ~\cite{You2020GraphCL}, MVGRL ~\cite{icml2020_1971}, and GRACE~\cite{Zhu:2020vf}.
Generally, these methods focus on devising different augmentation strategies to make the graph encoder robust to noise, typically by adding/removing nodes and edges from the original graph.  
Although these strategies achieve promising results to some extend, but they may offend the assumption that the positive augmentation views need to retain the same underlying label space with the original graph.
For example, as shown in Figure~\ref{motivation-visual}, dropping a carbon atom from the phenyl ring of aspirin breaks the molecular structure and forms an alkene chain in the molecular graph, and removing the edge between \textit{Tom} and \textit{Bob} would result in \textit{Tom} becoming an isolated node in the social graph.
Therefore, if the augmentation strategy is insufficient, it may generate inadequate graph views, which can easily mislead our encoder training and inevitably lead to over-confidence and over-fitting problems. 

Unlike previous GCL approaches that crudely determine that data augmentation does not change the labels of the graph/nodes, we want to generate some soft-confidence contrastive samples at the node level to enable fine-grained degrees of soft similarity between positive or negative pairs.
To achieve this idea, in this work, we propose a method called IDentity Mixup for Graph Contrastive Learning (ID-MixGCL). 
Since there is no ground-truth node label information available in the self-supervised setting, ID-MixGCL first assigns a distinct identity label to each node in the graph. 
Next, similar to previous GCL works, ID-MixGCL creates two augmented versions of a given original graph by randomly removing edges and masking attributes, and applies a shared GNN encoder to obtain node embeddings. 
The key difference is that ID-MixGCL linearly interpolates between two node embeddings in one of the augmented graphs, as well as performing the same linear interpolation on the associated pair of identity labels to adjust the label accordingly for contrastive loss.
This operation allows the simultaneous modulation of both input and label spaces with a controllable degree of change, leading to the capture of fine-grained representations from unlabelled graphs and the learning of more precise and smoother decision boundaries on latent features.

To evaluate the effectiveness of ID-MixGCL, we conduct a comprehensive series of experiments utilizing a diverse collection of datasets for both node classification and graph classification tasks. 
The datasets include 8 node classification datasets and 6 graph classification datasets. 
Our results for the node classification task indicate that our methodology surpasses the current state-of-the-art on all eight node-level datasets. 
Furthermore, for the graph classification task, our methodology demonstrates the best performance on five out of six datasets, with a relative improvement of up to 29.13\% compared to the previous top-performing baseline.

\section{Related Works}
\subsection{Data Augmentation}
\textbf{Input Augmentation} is a technique used to enhance the input graph by applying various strategies, such as attribute augmentation, node augmentation, edge augmentation, and graph augmentation. 
Attribute augmentation modifies the raw information of nodes in a graph, for example, by shuffling or masking attributes~\cite{Velickovic2019DeepGI,You2020GraphCL}. 
Node augmentation creates diverse data by manipulating graph nodes, such as generating new nodes or dropping existing ones~\cite{Feng2020GraphRN,Zhao2021GraphSMOTEIN}. 
Edge augmentation diversifies data by altering the topology of the graph through removing or adding edges~\cite{Rong2020DropEdgeTD,Zhao2020DataAF}. 
Graph augmentation modifies the overall structure of the graph, and can be seen as a combination of the previous methods~\cite{han2022g}. 
\textbf{Feature Augmentation}, on the other hand, aims at augmenting the samples in the feature space rather than the input graph. 
For example, by applying matrix-sketching augmentation on the hidden features or perturbing the GNN encoder~\cite{Zhang2022COSTACF}.
Research has shown that feature augmentation can achieve comparable performance to input augmentation, and that a combination of the two can lead to even better results~\cite{Xia2022SimGRACEAS,Gong2022MAGCLMA}, but it is important to note that both types of augmentation have not considered the potential changes in the underlying labels that may occur as a result of data augmentation. 
Our research aims to address this issue for the first time in the context of graph contrastive learning, and presents a method to effectively alleviate it, resulting in a significant performance gain.

\subsection{Graph Contrastive Learning}
Graph Contrastive Learning (GCL) is a technique that aims to create an embedding space in which similar graph pairs are close to each other, while dissimilar ones are far apart. Popular GCL approaches use a framework in which the original graph is augmented to generate two distinct views, and then the mutual information between corresponding objects in the two views is maximized. 
For example, GraphCL~\cite{You2020GraphCL} applies a series of graph augmentations on the original graph to generate an augmented graph, and learns whether two graphs are from the same graph. GRACE~\cite{Zhu:2020vf} uses both edge dropping and attribute masking to create different views of the graph, and maximize the mutual information of the same node between two different views of the original graph. 
MVGRL~\cite{icml2020_1971} applies graph diffusion kernels~\cite{Klicpera2019DiffusionIG} to the structure of the original graph and constructs graph views using a sampler that sub-samples identical nodes from both views. 
However, these methods assume that data augmentation does not lead to potential changes in the label space. 
This assumption is not always true, e.g. structural perturbations of drug molecules can lead to complete changes in the function and category of molecules. In this paper, we focus on this challenge for the first time, and propose an identity mixup strategy to soften the label space and alleviate the overfitting caused by the false-negative augmentations.

\begin{figure*}[ht]
   \centering
   \includegraphics[width=0.95\textwidth]{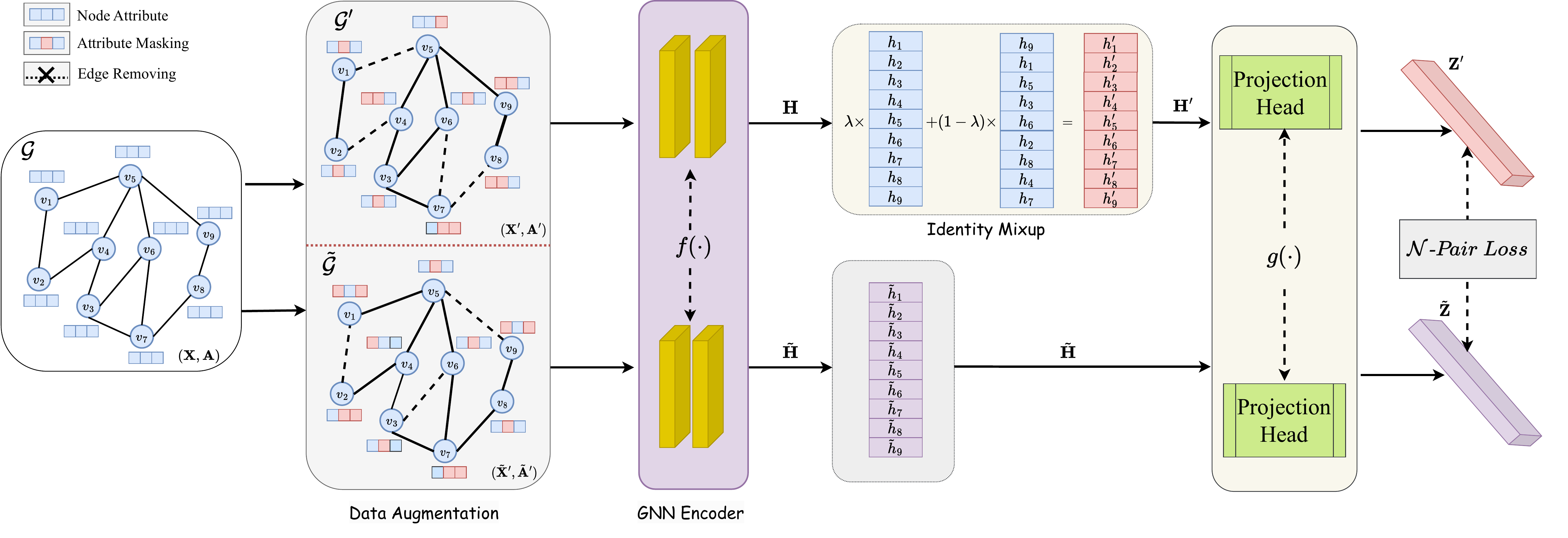}
   \caption{Overview of ID-MixGCL. The original graph $\mathcal{G}$ is used to augment two  different but semantically similar views, where $\mathcal{G}'=\{\bm{X}',\bm{A}'\}$ and $\tilde{\mathcal{G}}=\{\tilde{\bm{X}},\tilde{\bm{A}}\}$. After that, we feed the graph views $\mathcal{G}'$ and $\tilde{\mathcal{G}}$ into a GNN encoder $f(\cdot)$ and obtain two views of node representation matrix ($\bm{H}$ and $\tilde{\bm{H}}$). Then we make a mixup operation on the representation matrix $\bm{H}$, where $\bm{H}'$ is the mixed node representation matrix. After passing a shared projection head $g(\cdot)$, we use a contrastive loss to maximize the agreement between representations $\bm{Z}'$ and $\tilde{\bm{Z}}$.}
   \label{p-mixgcl}
\end{figure*}

\subsection{Mixup Training}
Mixup~\cite{zhang2018mixup} is a data augmentation technique that uses convex interpolations of input data and their corresponding labels to improve the generalisation capabilities of a model during training.
This method has been shown to be effective across a wide range of tasks, including supervised and semi-supervised graph representation learning. 
For instance, Wang~\cite{10.1145/3442381.3449796} proposed the use of a two-branch graph convolution for mixing receptive field subgraphs for paired nodes, which leverages the representations of each node's neighbors prior to the application of Mixup for graph convolutions. 
Additionally, Verma~\cite{Verma2020GraphMixIT} introduced GraphMix, a method that trains a fully-connected network in conjunction with a graph neural network through weight sharing and interpolation-based regularization, resulting in improved performance on semi-supervised object classification tasks. 
To the best of our knowledge, however, incorporating mixup into graph contrastive learning (GCL) has been a challenging task due to the lack of labels. 
In this work, we propose a novel approach to incorporate mixup into GCL by utilizing identity labels.
Our experimental results show that mixup has great potential in GCL. 

\section{methodology}
In this section, we will begin by introducing the relevant notation and then present our proposed method in five parts. The overall structure of our method is illustrated in Figure~\ref{p-mixgcl}.

\subsection{Notations}

Given a set of nodes $\{v_1,v_2,\dots,v_n\}$, a graph $\mathcal{G}=\{\bm{X},\bm{A}\}$ is defined by a node feature matrix $\bm{X}\subseteq\mathbb{R}^{n \times d}$ and an adjacency matrix $\bm{A}\in\{0,1\}^{n \times n}$, where each element $\bm{A}_{i,j}$ describes whether an edge exists nodes $v_i$ and $v_j$.
Additionally, we also introduce a diagonal degree matrix $\bm{D}$, where $\bm{D}_{i,i}$ is the number of edges incident with node $v_i$. 
The goal of GCL is to learn a graph encoder that maps the nodes to a low-dimensional dense vector space to fit downstream tasks.

\subsection{Graph Augmentation}

Several augmentation strategies has been successfully explored in GCL~\cite{Zhu:2020vf,You2020GraphCL}. 
In this work, we consider utilizing the two typical augmentation techniques to generate different views, including the attribute and edge augmentation:
\begin{align}
\small
    \bm{X}' &= \bm{X}\odot\bm{M}, \\
    \bm{A}' &= \bm{A}\odot\bm{R}, \\
    \mathcal{G}' &\triangleq\{\bm{X}', \bm{A}'\},
\end{align}
where $\bm{M}\in\{0,1\}^{n\times d}$ is an attribute masking matrix, whose entry independently is drawn from a $\textit{Bernoulli}$ distribution with probability $p_1$. $\bm{R}\in\{0,1\}^{n\times n}$ is a random edge masking matrix, it is also drawn from a $\textit{Bernoulli}$ distribution with probability $p_2$, $\odot$ is the Hadamard production. 
Here, we simultaneously conduct augmentations on the attributes and edges, and finally generate the augmented graph $\mathcal{G}'$.

\subsection{Graph Neural Network Encoder}
We use the well-known graph convolutional network (GCN) \cite{Kipf2017SemiSupervisedCW} as the encoder $f(\cdot)$ to obtain the node features. Each GCN layer can be defined as follows:
\begin{equation}
\small
   \begin{aligned}
       \bm{H}^{l+1} &= f(\bm{A},\bm{H}^{l};\bm{W}^{l})\\
       &=\sigma(\hat{\bm{D}}^{-\frac{1}{2}}\hat{\bm{A}}\hat{\bm{D}}^{-\frac{1}{2}}{\bm{H}^{l}}\bm{W}^{l}),
   \end{aligned}
\end{equation}
where $\bm{H}^{0}$ is the row node feature $\bm{X}$, 
$\hat{\bm{A}}=\bm{A}+\bm{I} $ is the adjacency matrix with added selfloops, $\hat{\bm{D}}$ is the degree matrix $\hat{D}_{ii}=\sum_j \hat{A}_{ij}$, $\sigma(\cdot)$ is the activation function, \eg, ReLU($\cdot$). 
In following sections, we omit the superscript of $\bm{H}^{l+1}$ as $\bm{H}$.

\subsection{Identity-labeling Mixup}
Let $(\bm{x}_i,\bm{y}_i)$ and $(\bm{x}_j,\bm{y}_j)$ be two training instances, where $\bm{x}_i$ and $\bm{x}_j$ refer to the input samples, $\bm{y}_i$ and $\bm{y}_j$ refer to their ground-truth label vectors, Mixup~\cite{zhang2018mixup}  produces the synthetic sample $(\tilde{\bm{x}},\tilde{\bm{y}})$  as follows:
\begin{align}
\small
    \lambda &\backsim \textit{Beta}(\alpha,\beta), \\
    \tilde{\bm{x}} &= \lambda \bm{x}_i + (1-\lambda) \bm{x}_j, \\
    \tilde{\bm{y}} &= \lambda \bm{y}_i + (1-\lambda) \bm{y}_j, 
\end{align}
where $\lambda\in [0,1]$ is a scalar mixing ratio sampled from a $\textit{Beta}(\alpha,\beta)$ distribution. 
Mixup extends the training distribution by incorporating the prior knowledge that  feature interpolation should lead to correlated label interpolation.
However, the above strategy is not straightforward for GCL, since there are no ground-truth node labels $y$ available in the self-supervised setting. 
To overcome this limitation, we propose utilizing the nodes' identity index as labels, and then we can generate a series of mixture data by mixing the node representations $\bm{H}$ and the identity labels.
Specifically, for each node $v_i$ in a graph $\mathcal{G}$, we first define a one-hot identity label vector $\bm{P}_i = \{0,\dots,1,\dots,0\}\in \{0,1\}^{n}$, where $n$ is the number of nodes in the graph, and $\bm{P}_{i,i} = 1$.
Intuitively, the identity label records the index of the node in a training batch $\mathcal{B}$, which helps the mixed node to locate the (two) original nodes before mixup, so as to encourage the mixed node feature to perceive the ``soft'' similarity between the mixed node and the original nodes in training loss (See in Eq.~(\ref{eq:loss})).
\begin{figure}[t]
   \centering
   \includegraphics[width=0.47\textwidth]{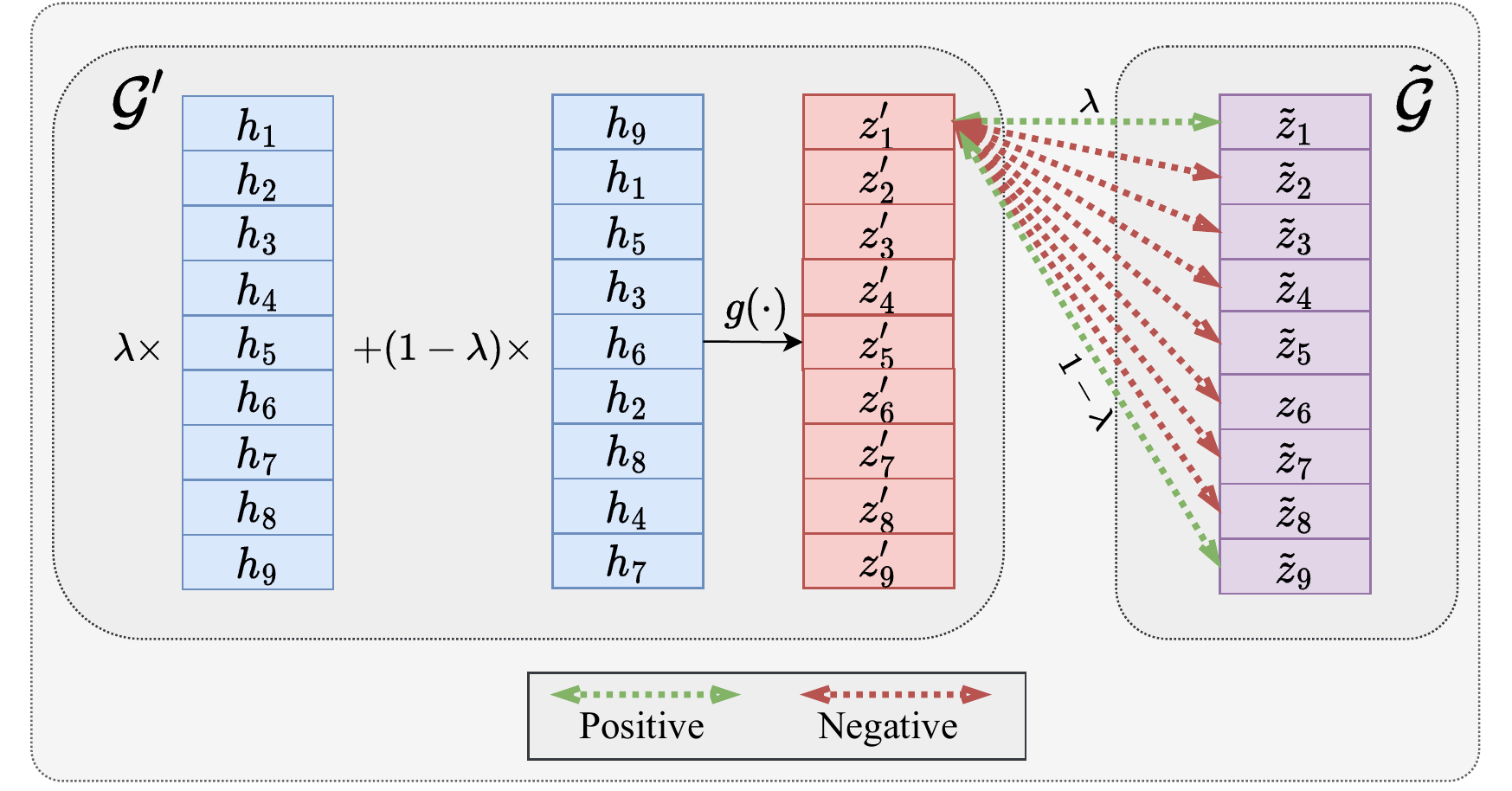}
   \caption{Illustration of our proposed identity mixup strategy. We select a different sample from the same batch and apply the mixup operation to it using a varied strategy for each sample. }
   \label{mixup-visual}
\end{figure}

In this work, we adapt three typical mixup strategies to achieve our GCL framework\footnote{Note that we focus on the mixup framework with identity-labels, and we select three well-studied mixup strategies for experiments. We will explore more newly-emerging mixup studies~\cite{Chen2021TransMixAT,Yang2022EnhancingCT} in future works.}:
\begin{itemize}
    \item \textbf{RandomMixup} is the vanilla mixup strategy, it randomly selects two nodes' embeddings for the mixup operation, which can be defined as follows:
        \begin{align}
        \small
           \bm{h}_j &= \texttt{Rand} (\mathcal{B}), \\
            \bm{h}_{i}' &= \lambda \bm{h}_i + (1-\lambda) \bm{h}_j,\\
            \bm{P}_{i}' &= \lambda \bm{P}_i + (1-\lambda) \bm{P}_j,
        \end{align}
    where $\texttt{Rand}(\mathcal{B})$ denotes a randomly selected node from the same batch $\mathcal{B}$.
    In this way, this strategy respectively mixes the features and the labels from two different nodes, and then encourages the model to focus on the common underlying features of the two nodes through contrastive learning.
    \item \textbf{CutMixup}~\cite{Yun2019CutMixRS} mixes paired samples with a random binary mask, which conducts the mixup strategy at the element-level. Formally, it can be defined as follows:
        \begin{align}
        \small
            \bm{h}_j &= \texttt{Rand} (\mathcal{B}), \\
            \bm{h}_{i}' &= \bm{m}_\lambda \odot \bm{h}_i + (\mathbbm{1}-\bm{m}_\lambda) \odot \bm{h}_j,\\
            \bm{P}_{i}' &= \lambda \bm{P}_i + (1-\lambda) \bm{P}_j,
        \end{align}
    where $\bm{m}_\lambda\in\{0,1\}^d$ indicates a mask binary vector from a $\textit{Bernoulli}$ distribution with probability $\lambda$, $\mathbbm{1}$ is a vector of ones, and $\odot$ is an element-wise multiplication.
    In this way, this strategy conducts mixup at the element-level (instead of vector-level in RandomMixup), providing more multifarious mixed features for learning robustness.
    \item \textbf{LocalMixup}~\cite{Baena2022PreventingMI} selects nodes with the closest geometric distance, and generate the mixed features as: 
        \begin{align}
        \small
            \exists \bm{h}_j &\in \mathcal{B}, \texttt{Dis}(\bm{h}_i,\bm{h}_j)\to 0, \\
            \bm{h}_{i}' &= \lambda \bm{h}_i + (1-\lambda) \bm{h}_j,\\
            \bm{P}_{i}' &= \lambda \bm{P}_i + (1-\lambda) \bm{P}_j,
        \end{align}
    where $\texttt{Dis}(\bm{h}_i,\bm{h}_j)$  denotes the L2 distance to find the closest node candidate $j$ to the central node $i$.
    In this way, this strategy tends to select the two nodes with high similarity, leading to high-quality mixed features with few out-of-distribution features or contradictory features~\cite{Baena2022PreventingMI}.
\end{itemize}

\begin{algorithm}[t]
	\caption{Pseudocode of ID-MixGCL With `RandomMixup' Strategy in Pytorch-like Style.}
	\label{alg:code}
	\definecolor{codeblue}{rgb}{0.25,0.5,0.5}
	\lstset{
		backgroundcolor=\color{white},
		basicstyle=\fontsize{7.2pt}{7.2pt}\ttfamily\selectfont,
		columns=fullflexible,
		breaklines=true,
		captionpos=b,
		commentstyle=\fontsize{7.2pt}{7.2pt}\color{codeblue},
		keywordstyle=\fontsize{7.2pt}{7.2pt},
	}
	\begin{lstlisting}[language=python]
# model: 2-layers GCN
# project: 2-layers MLP
# temp: temperature
for X,A in loader:  # load a batch x with N samples
    G1 = Aug1(X,A)   # Aug1 is a random augmentation
    G2 = Aug2(X,A)  # Aug2 is a another random augmentation
    
    # Get embeddings: NxF
    h1 = model(G1)
    h2 = model(G2)
    
    # Initialize the mixing ratio
    lam = np.random.beta(alpha, alpha)
    lam = max(lam, 1-lam)

    # Initialize the identity labels of nodes in the same batch
    batch_size = h1.size()[0]
    id_label = arange(batch_size)

    # Mixup Type: RandomMixup 
    mixed_id = torch.randperm(batch_size)
    mixed_h1 = lam * h1 + (1 - lam) * h1[mixed_id, :]
    
    # Projection Head
    mixed_z1, z2 = [project(x) for x in [mixed_h1, h2]]

    # Loss Function
    pred = torch.div(torch.matmul(mixed_z1, z2.t()), temp)
    loss = lam * nn.CrossEntropyLoss(pred, id_label) +\
            (1-lam) * nn.CrossEntropyLoss(pred, mixed_id)

    # Adam update: Network
    loss.backward()
    update(model.params)
	\end{lstlisting}
\end{algorithm}

\subsection{Projection Head}
Previous researches employ multi-layer perceptrons (MLPs) with hidden layers to improve the performance of downstream tasks, such as node classification. Our study also utilizes a two-layer MLP to obtain $\bm{Z}'$ and $\tilde{\bm{Z}}$.
\begin{align}
\small
    \bm{Z}' &= g(\bm{H}') = \bm{W}^{(2)} \sigma (\bm{W}^{(1)} \bm{H}'),\\
     \tilde{\bm{Z}} &= g(\tilde{\bm{H}}) = \bm{W}^{(2)} \sigma (\bm{W}^{(1)} \tilde{\bm{H}}),
\end{align}
where $g(\cdot)$ represents a projection head, and $\bm{W}^{(1)}$ and $\bm{W}^{(2)}$ are the weights of the hidden layers within the projection head, $\bm{H}'$ denotes the mixup node representations and $\tilde{\bm{H}}$ denotes the encoder output of another graph view.

\subsection{Loss Function}

In GCL, many previous studies have employed the InfoNCE~\cite{Zhang2022COSTACF} contrastive loss, which aims to maximize the similarity of positive pairs and minimize that of negative pairs. 
In mixup setting, each mixture feature $\bm{z}'$ has two positive sample and $(N-2)$ negative samples with $\tilde{\bm{Z}}$. 
We thereby introduce the $\mathcal{N}\textit{-Pair}$ contrastive loss~\cite{NIPS2016_6b180037} for $i$-th node as follows:
\begin{equation} \label{eq:loss}
\small
    \begin{aligned}
        \mathcal{L}_{\textit{Mix}} &=-\sum_{i=1}^N\sum_{j=1}^N \big[ \bm{P}_{i,j}'\log\frac{\exp{(\texttt{sim}(\bm{z}_{i}',\tilde{\bm{z}}_j)/\tau)}}{\sum_{k=1}^N\exp{(\texttt{sim}(\bm{z}_{i}',\tilde{\bm{z}}_k)/\tau)}}\big],       
    \end{aligned}
\end{equation}
Some details are illustrated in Figure~\ref{mixup-visual} for better understanding (note that, for each node $i$, our mixup label vector $\bm{P}_i'$ only two dimensions are positive, i.e. $\bm{P}_{i,i}'$ and $\bm{P}_{i,j}'$). 
In detail, $\bm{z}_i'=g(\lambda \bm{h}_i + (1-\lambda) \bm{h}_j)$ and $\tilde{\bm{z}}_i=g(\tilde{\bm{h}}_i)$ are obtained from two generated graph views $\mathcal{G}'$ and $\tilde{\mathcal{G}}$. 
For a mixed sample $\bm{z}'_i$, we now have two positive examples, $\tilde{\bm{z}}_i$ and $\tilde{\bm{z}}_j$, with desired similarities of $\lambda$ (e.g., $\bm{P}_{i,i}'$) and $1-\lambda$ (e.g., $\bm{P}_{i,j}'$), respectively.
This approach uses weighting to adjust the loss values, which is equivalent to scaling the similarity distance of positive and negative pairs in label space. 
This softens the label space and allows the model to learn a more fine-grained and robust representation. For expedient comprehension, we proffer the pseudocode of the RandomMixup strategy, see Algorithm \ref{alg:code}.

\section{Experiments}
In this section, we evaluate the performance of ID-MixGCL on both graph-level and node-level tasks, comparing them to other state-of-the-art methods. We also provide further analysis on the property of our proposed model.  
\subsection{Datasets}
\begin{table}[t]
\footnotesize
    \centering
    \caption{Datasets statistics for node classification.}
    \resizebox{\linewidth}{!}{\begin{tabular}{cccccc}
    \toprule
        \textsc{Dataset} & \textsc{Category}&  \textsc{Nodes} &  \textsc{Edges} &  \textsc{Features} &  \textsc{Classes} \\ \midrule
        Cora & \textsc{Citation Networks} & 2,708 & 5,429 & 1,433 & 7 \\ 
        Citeseer & \textsc{Citation Networks} &  3,327 & 4,732 & 3,703 & 6 \\ 
        Pubmed & \textsc{Citation Networks} &  19,717 & 44,338 & 500 & 3 \\
        DBLP & \textsc{Citation Networks} & 17,716 & 105,734 & 1,639 & 4 \\ \midrule
        Coauthor-CS & \textsc{Cooperative Networks} & 18,333 & 81,894 & 6,805 & 15 \\
        Coauthor-Physics & \textsc{Cooperative Networks} & 34,493 & 247,962 & 8,415 & 5 \\ \midrule
        Amazon-Computers & \textsc{Copurchase Networks} & 13,752 & 245,861 & 767 & 10 \\
        Amazon-Photo & \textsc{Copurchase Networks} & 7,650 & 119,081 & 745 & 8 \\ \bottomrule
    \end{tabular}}
    \label{node-classify-datasets}
\end{table}
\begin{table}[t]
\footnotesize
    \centering
    \caption{Datasets statistics for graph classification.}
    \resizebox{\linewidth}{!}{\begin{tabular}{ccccccc}
    \toprule
         \textsc{Datasets} &  \textsc{Category}  & \textsc{Graph} & \textsc{Avg. Nodes} &  \textsc{Avg. Edges} &  \textsc{Classes} \\ \midrule
        IMDB-B & \textsc{Social Networks} &1000 & 19.8 & 193.1 & 2 \\ 
        IMDB-M & \textsc{Social Networks}& 321 & 22.35 & 249.46 & 3 \\ 
        COLLAB & \textsc{Social Networks}& 5,000 & 74.5 & 4914.4 & 3 \\ \midrule
        MUTAG & \textsc{Biochemical Molecules} & 188 & 17.9 & 39.6 & 2 \\
        NCI1 &\textsc{Biochemical Molecules}&4110 & 29.84 & 32.37 & 2 \\ 
        PROTEINS &\textsc{Biochemical Molecules} & 1,113 & 39.1 & 145.6 & 2 \\ \bottomrule
    \end{tabular}}
    \label{graph-classify-datasets}
\end{table}
\subsubsection{Node Classification Datasets}
The datasets utilized in node classification task can be see in Table \ref{node-classify-datasets}, These datasets are described as follows:
(a) Citation networks including Cora, Citeseer, Pubmed, and DBLP datasets~\cite{PrithvirajSen2008CollectiveCI}, are constructed based on citation relationships in specific research fields.
(b) Copurchase networks including Amazon Computers and Amazon Photo~\cite{JulianMcAuley2015ImagebasedRO}, are constructed based on co-purchase relationships extracted from Amazon. 
(c) Cooperative networks include coauthor-CS and coauthor-physics datasets~\cite{ArnabSinha2015AnOO}, which consist of co-authorship graphs derived from the Microsoft Academic Graph Dataset. 
\subsubsection{Graph Classification Datasets}
The datasets utilized in graph classification task can be see in Table \ref{graph-classify-datasets}, These datasets are described as follows:
(a) Biochemical molecule include PROTEINS~\cite{KarstenMBorgwardt2005ProteinFP}, MUTAG~\cite{AsimKumarDebnath1991StructureactivityRO}, and NCI1~\cite{Wale2006ComparisonOD}. 
These datasets are commonly used for tasks such as molecular classification and protein function prediction.
(b) Social networks, derived from real-world social activity, include the IMDB-B, IMDB-M~\cite{PinarYanardag2015DeepGK}, and COLLAB~\cite{JureLeskovec2005GraphsOT} datasets. These datasets are utilized in the study of network analysis and prediction tasks.
\subsection{Experimental Setting}
\subsubsection{Computing infrastructures}
All models were implemented using PyTorch Geometric 2.1.0 and PyTorch 1.10.1. All datasets used through the experiment are benchmark datasets provided by PyTorch Geometric libraries. All experiments were conducted on a computer server with four NVIDIA Tesla V100 GPUs (each with 32 GB of RAM) and fourteen Intel Xeon E5-2660 v4 CPUs.

\subsubsection{Evaluation Protocols}
Following previous GCL methods~\cite{You2020GraphCL,Zhu:2020vf}, we follow the paradigm of \textit{Pre-training + Fine-tuning}. (1) \textit{Self-supervised pre-training}: ID-MixGCL learns the node embeddings through self-supervised training, and obtains the graph embeddings by pooling all node embeddings in a graph. (2) \textit{Fine-tune the downstream tasks}: For the node classification task, we use the node embeddings learned by pre-training, fine-tune on the training sets, and use a simple logistic classifier with $L_2$ regularization to obtain the results on the test sets. For the graph classification task, we pool the node embeddings to obtain the graph embeddings, and then feed the graph embeddings into the downstream SVM classifier with 10-fold cross-validation.

\subsubsection{Compared Baselines}
For the \textit{node classification task}, We selected three representative benchmark methods for comparison: random walk-based models (DeepWalk~\cite{BryanPerozzi2014DeepWalkOL} and node2vec~\cite{AdityaGrover2016node2vecSF}), autoencoder-based models (GAE and VGAE~\cite{ThomasNKipf2022VariationalGA}), and contrastive-based models (DGI~\cite{Velickovic2019DeepGI}, GRACE~\cite{Zhu:2020vf}, GCA~\cite{YanqiaoZhu2020GraphCL}, and COSTA~\cite{Zhang2022COSTACF}). 
For the \textit{graph classification task}, to fairly and comprehensively evaluate our approach, we used three categories of baselines:
kernel-based methods, such as the Weisfeiler-Lehman sub-tree kernel (WL)~\cite{NinoShervashidze2011WeisfeilerLehmanGK} and deep graph kernels (DGK)~\cite{PinarYanardag2015DeepGK}, classic unsupervised methods such as node2vec ~\cite{BalajiKrishnapuram2016ProceedingsOT}, sub2vec~\cite{BijayaAdhikari2018Sub2VecF}, and graph2vec~\cite{AnnamalaiNarayanan2017graph2vecLD}, and the self-supervised learning methods such as Infograph~\cite{FanYunSun2019InfoGraphUA}, GraphCL~\cite{You2020GraphCL}, JOAO~\cite{You2021GraphCL}, MVGRL~\cite{icml2020_1971}, AutoGCL~\cite{YihangYin2021AutoGCLAG}, and SimGRACE~\cite{JunXia2022SimGRACEAS}. In particular, raw features were directly used to conduct the node classification task.
The experimental results for all baselines are reported from their published papers to avoid re-implementation bias.

\subsubsection{Implementation and Hyperparameter Setting}
All experiments are are initialized with Glorot initialization~\cite{XavierGlorot2010UnderstandingTD}, and trained using Adam SGD optimizer~\cite{DiederikPKingma2014AdamAM}. 
The $l_2$ weight decay factor is set to $10^{-5}$ on all datasets. 
We use $p_{e,1},p_{e,2},p_{f1},p_{f2}$ as the the probability parameters that control the sampling process, where $p_{e,1},p_{f1}$ for the one augmented view and $p_{e,2},p_{f2}$ for the another augmented view. $p_{e,1},p_{e,2}$, is used for controlling the ratio of dropping edges and $p_{f1},p_{f2}$ decides what a fraction of attribute dimensions will be masked. 
The probability parameters $p_{e,1},p_{e,2},p_{f1},p_{f2}$ is selected from $\{0.0, 0.1$, 0.2, 0.3, 0.4 $0.5\}$,
the activation function is selected from \{ReLU, LeakyReLU, PReLU, RReLU\}, the learning rate is selected from $\{0.01, 0.001$, 0.0005, $0.0001\}$, the batch size is selected from $\{128$, 256, $512\}$, the temperature is selected from $\{0.2, 0.4\}$, and we fixed sampling of $\lambda$ from $ \textit{Beta}(1,1)$. 
Furthermore, a fixed architecture has been adopted, which comprises of utilizing 2 layers of GCN as encoder, and 2 layers of MLP as the projection head. 
\begin{table*}[ht]
\footnotesize
    \centering
    \caption{Summary of the accuracy (\%) (± std) on node classification, where $X$ denotes the initial node features, $A$ denotes the djacency matrix. The highest performance of unsuperviced models is highlighted in boldface.}
    \setlength{\tabcolsep}{21pt}{
    \begin{tabular}{lccccc}
    \toprule
        \textbf{Method} & \textbf{Training Data} & \textbf{Cora} & \textbf{Citeseer} & \textbf{Pubmed} & \textbf{DBLP}  \\ 
        \midrule
        Raw features & \textbf{X} & 64.8 & 64.6 & 84.8 & 71.6 \\ 
        node2vec~\cite{AdityaGrover2016node2vecSF} & \textbf{A} & 74.8 & 52.3 & 80.3 & 78.8  \\ 
        DeepWalk~\cite{BryanPerozzi2014DeepWalkOL} & \textbf{A} & 75.7 & 50.5 & 80.5 & 75.9  \\ 
        \midrule
        DeepWalk~\cite{BryanPerozzi2014DeepWalkOL} & \textbf{X,A} & 73.1 & 47.6 & 83.7 & 78.1  \\ 
        GAE~\cite{ThomasNKipf2022VariationalGA} & \textbf{X,A} & $76.9\pm 0.0$ & $60.6\pm 0.0$ & $82.9\pm 0.0$ & $81.2\pm 0.0$  \\ 
        VAGE~\cite{ThomasNKipf2022VariationalGA} & \textbf{X,A} & $78.9\pm 0.0$ & $61.2\pm 0.0$ & $83.0\pm 0.0$ & $81.7\pm 0.0$  \\ 
        DGI~\cite{Velickovic2019DeepGI} & \textbf{X,A} & $82.6\pm 0.4$ & $68.8\pm 0.7$ & $86.0\pm 0.1$ & $83.2\pm 0.1$  \\ 
        GRACE~\cite{Zhu:2020vf} & \textbf{X,A} & $83.3\pm 0.4$ & $72.1\pm 0.5$ & $86.7\pm 0.1$ & $84.2\pm 0.1$ \\ 
        GCA~\cite{YanqiaoZhu2020GraphCL} & \textbf{X,A} & $82.8\pm 0.3$ & $71.5\pm 0.3$ & $86.0\pm 0.2$ & $83.1\pm 0.2$ \\ 
        COSTA~\cite{Zhang2022COSTACF} & \textbf{X,A} & $84.3\pm 0.2$ & $72.9\pm 0.3$ & $86.0\pm 0.2$ & $84.5\pm 0.1$  \\ 
        \midrule
        ID-MixGCL & \textbf{X,A} & $\mathbf{87.1\pm 0.1}$ & $\mathbf{75.4\pm 0.2}$ & $\mathbf{88.2\pm 0.1}$ & $\mathbf{86.1\pm 0.1}$ \\ 
        \bottomrule
    \end{tabular}}
    \label{node-classify-1}
\end{table*}
\begin{table*}[ht]
\footnotesize
    \centering
    \caption{Summary of the accuracy (\%) (± std) on  node classification, and we use the same settings in Table \ref{node-classify-1}.}
    \setlength{\tabcolsep}{16pt}{
    \begin{tabular}{lccccc}
    \toprule
        \textbf{Method} & \textbf{Training Data} & \textbf{Coauthor-CS} & \textbf{Coauthor-Phy} & \textbf{Amazon-Photo} & \textbf{Amazon-Computers} \\ 
        \midrule
        Raw features & \textbf{X} & 90.37 & 93.58 & 78.53 & 73.81 \\ 
        node2vec~\cite{AdityaGrover2016node2vecSF} & \textbf{A} & 85.08 & 91.19 & 89.67 & 84.39 \\ 
        DeepWalk~\cite{BryanPerozzi2014DeepWalkOL} & \textbf{A} & 84.61 & 91.77 & 89.44 & 85.68 \\ 
        \midrule
        DeepWalk~\cite{BryanPerozzi2014DeepWalkOL} & \textbf{X,A} & 87.7 & 94.9 & 90.05 & 86.28 \\ 
        GAE~\cite{ThomasNKipf2022VariationalGA} & \textbf{X,A} & $90.01\pm 0.71$ & $94.92\pm 0.07$ & $91.62\pm 0.13$ & $85.27\pm 0.19$ \\ 
        VAGE~\cite{ThomasNKipf2022VariationalGA} & \textbf{X,A} & $92.11\pm 0.00$ & $94.52\pm 0.00$ & $92.2\pm 0.11$ & $86.37\pm 0.21$ \\ 
        DGI~\cite{Velickovic2019DeepGI} & \textbf{X,A} & $92.15\pm 0.63$ & $94.51\pm 0.52$ & $91.61\pm 0.22$ & $83.95\pm 0.47$ \\ 
        MVGRL~\cite{icml2020_1971} & \textbf{X,A} & $92.11 \pm 0.12$ & $95.33\pm 0.03$ & $91.74\pm 0.07$ & $87.52\pm 0.11$\\
        GRACE~\cite{Zhu:2020vf} & \textbf{X,A}  & $92.95\pm 0.03$ & $95.72\pm 0.03$ & $92.53\pm 0.16$ & $87.80\pm 0.23$ \\ 
        GCA~\cite{YanqiaoZhu2020GraphCL} & \textbf{X,A} & $92.95\pm 0.13$ & $95.73\pm 0.03$ & $92.24\pm 0.21$ & $87.54\pm 0.49$ \\ 
        COSTA~\cite{Zhang2022COSTACF} & \textbf{X,A} & $92.94\pm 0.10$ & $95.60\pm 0.02 $& $92.56\pm 0.45$ & $88.32\pm 0.03$ \\ 
        \midrule
        ID-MixGCL & \textbf{X,A} & $\mathbf{93.50 \pm 0.11}$ & $\mathbf{96.20\pm 0.03}$ & $\mathbf{95.69\pm 0.02}$  & $\mathbf{88.40\pm 0.02}$ \\ 
        \bottomrule
    \end{tabular}}
    \label{node-classify-2}
\end{table*}
\begin{table*}[!h]
\footnotesize
    \centering
    \caption{Graph classification accuracy (\%) with standard deviation on TU datasets \cite{Ivanov2019UnderstandingIB}. 
    We report the mean 10-fold cross-validation accuracy over 5 runs.
    The highest performance of unsupervised models is highlighted in boldface. $-$ indicates that results are not available in published papers.}
    \setlength{\tabcolsep}{14pt}{
    \begin{tabular}{lcccccc}
    \toprule
        \textbf{Method} & \textbf{IMDB-B} & \textbf{IMDB-M} & \textbf{COLLAB} & \textbf{PROTEINS} & \textbf{MUTAG} & \textbf{NCI1} \\ 
        \midrule
        WL~\cite{NinoShervashidze2011WeisfeilerLehmanGK} & 72.30 ± 3.44 & 46.95 ± 0.46 & - & 72.92 ± 0.56 & 80.72 ± 3.00 & 80.31 ± 0.46 \\ 
        DGK~\cite{PinarYanardag2015DeepGK} & 66.96 ± 0.56 & 44.55 ± 0.52 & - & 73.30 ± 0.82 & 87.44 ± 2.72 & 80.31 ± 0.46 \\ 
        \midrule
        node2vec~\cite{BalajiKrishnapuram2016ProceedingsOT} & 50.2 ± 0.9 & 36.0 ± 0.7 & - & 57.49 ± 3.57 & 72.63 ± 10.20 & 54.89 ± 1.61 \\ 
        sub2Vec~\cite{BijayaAdhikari2018Sub2VecF} & 55.26 ± 1.54 & 36.7 ± 0.8 & - & 53.03 ± 5.55 & 61.05 ± 15.80 & 52.84 ± 1.47 \\ 
        graph2vec~\cite{AnnamalaiNarayanan2017graph2vecLD} & 71.10 ± 0.54 & 50.44 ± 0.87 & - & 73.30 ± 2.05 & 83.15 ± 9.25 & 73.22 ± 1.81 \\ 
        \midrule
        Infograph~\cite{FanYunSun2019InfoGraphUA} & 73.03 ± 0.87 & 49.69 ± 0.53 & 70.65 ± 1.13 & 74.44 ± 0.31 & 89.01 ± 1.13 & 76.20 ± 1.06 \\ 
        GraphCL~\cite{You2020GraphCL} & 71.14 ± 0.44 & 48.58 ± 0.67 & 71.36 ± 1.15 & 74.39 ± 0.45 & 86.80 ± 1.34 & 77.87 ± 0.41 \\ 
        JOAO~\cite{You2021GraphCL} & 70.21 ± 3.08 & 49.20 ± 0.77 & 69.50 ± 0.36 & 74.55 ± 0.41 & 87.35 ± 1.02 & 78.07 ± 0.47 \\ 
        MVGRL~\cite{icml2020_1971} & 74.20 ± 0.70 & 51.20 ± 0.50 & - & 71.5 ± 0.30 & \textbf{89.70 ± 1.10} & 75.1 ± 0.50 \\ 
        AutoGCL~\cite{YihangYin2021AutoGCLAG} & 73.30 ± 0.40 & - & 70.12 ± 0.68 & 75.80 ± 0.36 & 88.64 ± 1.08 & 82.00 ± 0.29 \\ 
        SimGRACE~\cite{JunXia2022SimGRACEAS} & 71.30 ± 0.77 & - & 71.72 ± 0.82 & 75.35 ± 0.09 & 89.01 ± 1.31 & 79.12 ± 0.44 \\ 
        \midrule
        ID-MixGCL & \textbf{98.53 ± 0.36} & \textbf{72.25 ± 3.30} & \textbf{79.85 ± 0.68} & \textbf{89.95 ± 2.20} & 89.25 ± 0.80 & \textbf{84.41± 1.60} \\ 
        \bottomrule
    \end{tabular}}
    \label{Grpah-classify-result}
\end{table*}
\subsection{Downstream Task Evaluation}
\subsubsection{Node Classification}\label{node—classification}
In line with previous researches~\cite{Zhu:2020vf,Zhang2022COSTACF}, our model was trained in an unsupervised manner. The embeddings obtained in this step were used to train and test a simple logistic classifier with $L_2$ regularization. The classifier was trained for 20 runs and the mean accuracy was used to measure performance on node classification tasks. Our results, displayed in Table \ref{node-classify-1} and Table \ref{node-classify-2}, indicate that our method outperforms current state-of-the-art methods on all eight node-level datasets. Our experiments demonstrate superior performance on all three types of networks (citation, cooperation, and co-purchase) across eight datasets in comparison to other existing approaches.
\begin{table}[ht]
\footnotesize
    \centering
    \caption{The performance comparison with mixup ratio $\lambda$. }
    \resizebox{\linewidth}{!}{\begin{tabular}{lcccc}
    \toprule
        & Cora & Citeseer& IMDB-M & PROTEINS \\ \cmidrule(r){2-2} \cmidrule(r){3-3} \cmidrule(r){4-4} \cmidrule(r){5-5} 
        & Accuracy(\%)  & Accuracy(\%)  & Accuracy(\%) & Accuracy(\%) \\ 
        \midrule
        $\lambda = 0.1$ & 83.52 ± 1.02 & 72.63 ± 1.29 & 64.60 ± 0.41 & 87.27± 0.34 \\ 
        $\lambda = 0.3$& 83.82 ± 1.13 & 74.73 ± 0.62 & 67.37 ± 0.66 & 87.56 ± 0.38 \\ 
        $\lambda = 0.5 $ & 86.47 ± 1.53 & 75.02 ± 0.55 & 71.62 ± 0.67 & 90.13 ± 1.37 \\ 
        \bottomrule
    \end{tabular}}
    \label{lambda}
\end{table}
\begin{figure}[ht]
  \centering
  \includegraphics[width=0.45\textwidth]{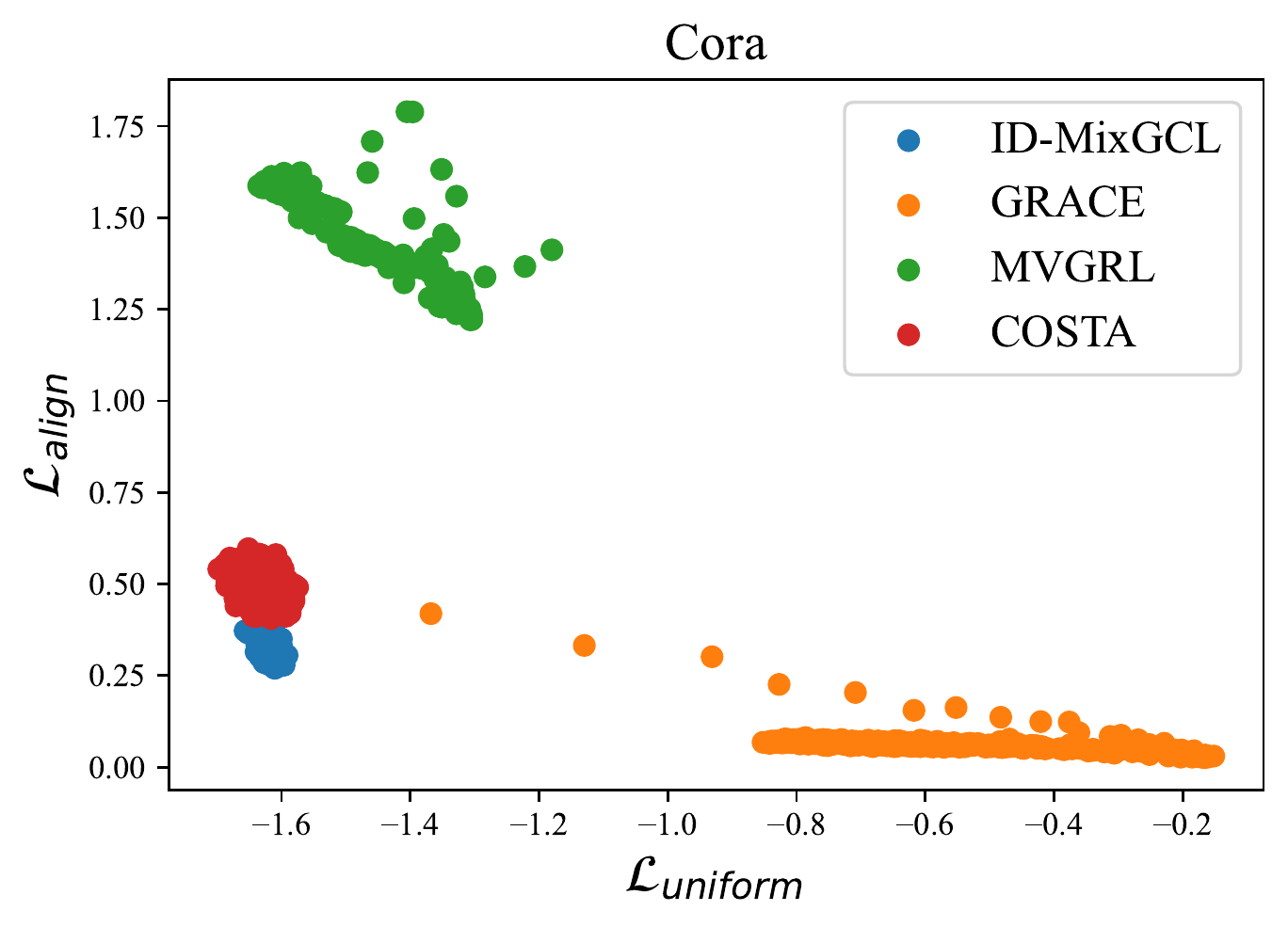}  \caption{$\mathcal{L}_{\texttt{align}}$-$\mathcal{L}_{\texttt{uniform}}$ plot of ID-MixGCL,GRACE, MVGRL and COSTA on Cora dataset. For both  metrics, lower is better.}
  \label{align-uniform-visual}
\end{figure}
GRACE can be regarded as the basic version of our method without the identity mixup strategy.
Compared with GRACE, our method has improved on all the datasets, with an average increase of 1.9pts. 
We attribute this improvement to the addition of identity mixing, which improves the diversity of the samples.

\subsubsection{Graph Classification}\label{graph—classification}
In order to ensure a fair and unbiased assessment of the performance of our method, we used the consistent evaluation functions of GraphCL and SimGRACE. By doing so, we ensured that our evaluations were conducted using standardized and widely accepted metrics. To evaluate our method, we used a 10-fold cross-validation procedure, reporting the mean accuracy with standard deviation after 5 runs. We also used a downstream support vector machine (SVM) classifier to train the classifier. The results, as shown in Table \ref{Grpah-classify-result}, illustrate that our approach achieved state-of-the-art results among self-supervised models.
Specifically, we significantly improve performance on various graph-level datasets, particularly the IMDB-B, IMDB-M, and PROTEINS datasets, where it outperforms the current state-of-the-art methods by a margin of more than 10pts. 
In graph classification tasks, GraphCL is a strong baseline for ID-MixGCL. 
Our method can be considered as adding identity mixup on top of it. 
This operation brings an average of 11.1pts improvement on 5 datasets, also proving the effectiveness of our proposed method.

We notice that the improvement in graph classification tasks is more significant than in node classification (11.1pts vs 1.9pts), and we suspect that this is because structurally augmenting the graph can easily change the graph's class labels, causing previous GCL methods to force two augmented graphs with different labels to be similar, which inevitably introduces bias. 
On the node level, the node label changes caused by disturbance to the graph structure are relatively infrequent, so previous GCL methods worked to some extent. 
Our ID-MixGCL avoids the introduction of bias by relaxing the previous assumption: operating on input features and labels at the same time, thus allowing the model to learn more robust node representations.

\subsection{Detailed Analysis}
In this section, we aim to address six research questions: 
\begin{itemize}
    \item \textbf{Q1}, whether mixing up input features and label space simultaneously will lead to improved performance (See in \S~\ref{sec:q1})?
    \item \textbf{Q2}, whether our method learns a better representation by softening the label space (See in \S~\ref{sec:q2})?
    \item \textbf{Q3}, which of the three Mixup strategies is better (See in \S~\ref{sec:q3})?
    \item \textbf{Q4}, can the better and more robust representation we have learned bring other benefits (See in \S~\ref{sec:q4})?
    \item \textbf{Q5}, what's the impact of multiple graph views on the results (See in \S~\ref{sec:q5})?
    \item \textbf{Q6}, how well does the representation learned by our model compared to other model representations via t-SNE visualization (See in \S~\ref{sec:q6})?
\end{itemize}

\subsubsection{Performance w.r.t $\lambda$ (Q1)}\label{sec:q1}
In sections \S~\ref{node—classification} and \S~\ref{graph—classification}, experimental results have shown that incorporating identity mixup can significantly improve performance on node and graph classification tasks. 
A natural question that arises is whether the higher the mixup ratio, the greater the change in the feature and label space, and the more obvious the effect improvement? 
We conduct experiments on four node and graph classification datasets by selecting three values of mixup's $\lambda$, 0.1, 0.3, 0.5 (note that because the mixup operation and loss are symmetric, the range of lambda values is [0, 0.5]). 
As shown in Table \ref{lambda}, we observe on all four datasets that as $\lambda$ values approach 0.5, the performance is better. 
This illustrates the value of softening the feature space and label space: the higher the mixup ratio, the softer the label space, the more it can encourage the model to learn alignment between fine-grained positives, making the model less affected by false positives caused by data augmentation. 

\begin{figure*}[htbp]
    \centering
    \subfigure[Cora]{
    \begin{minipage}[t]{0.25\linewidth}
    \centering
    \includegraphics[width=1.0\linewidth]{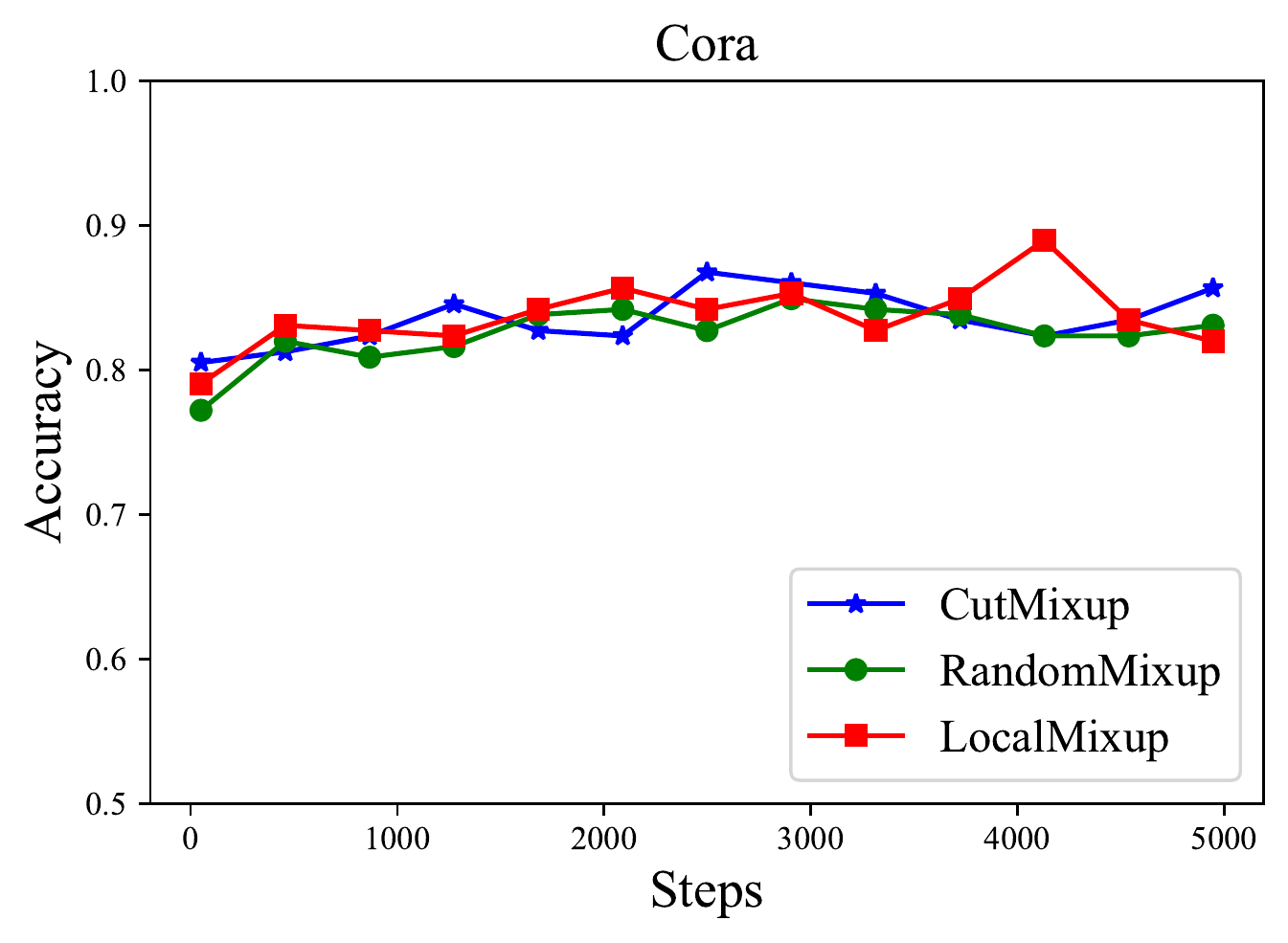}
    \end{minipage}%
    }%
    \subfigure[Citeseer]{
    \begin{minipage}[t]{0.24\linewidth}
    \centering
    \includegraphics[width=1.0\linewidth]{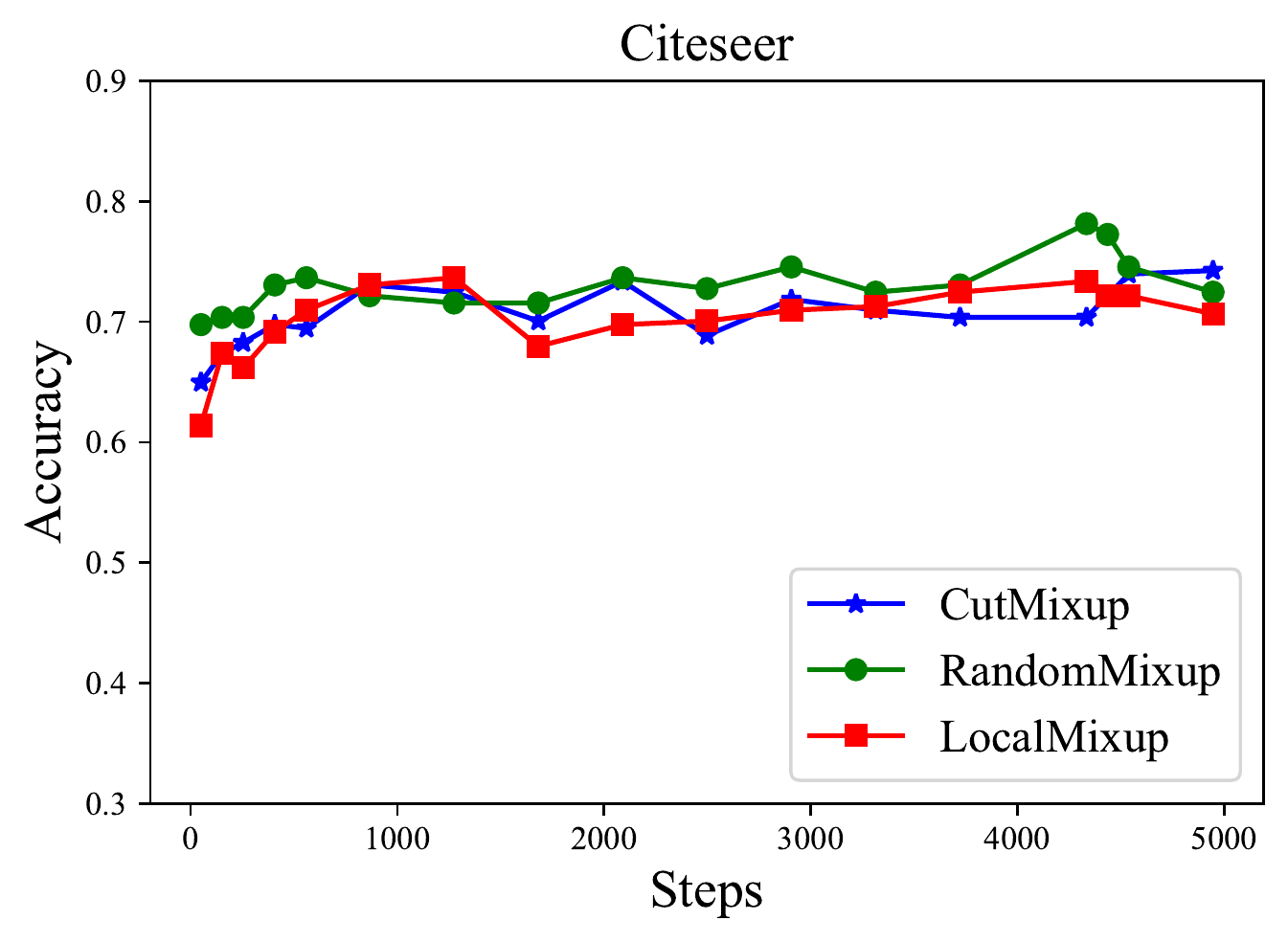}
    \end{minipage}%
    }%
    \subfigure[IMDB-M]{
    \begin{minipage}[t]{0.24\linewidth}
    \centering
    \includegraphics[width=1.0\linewidth]{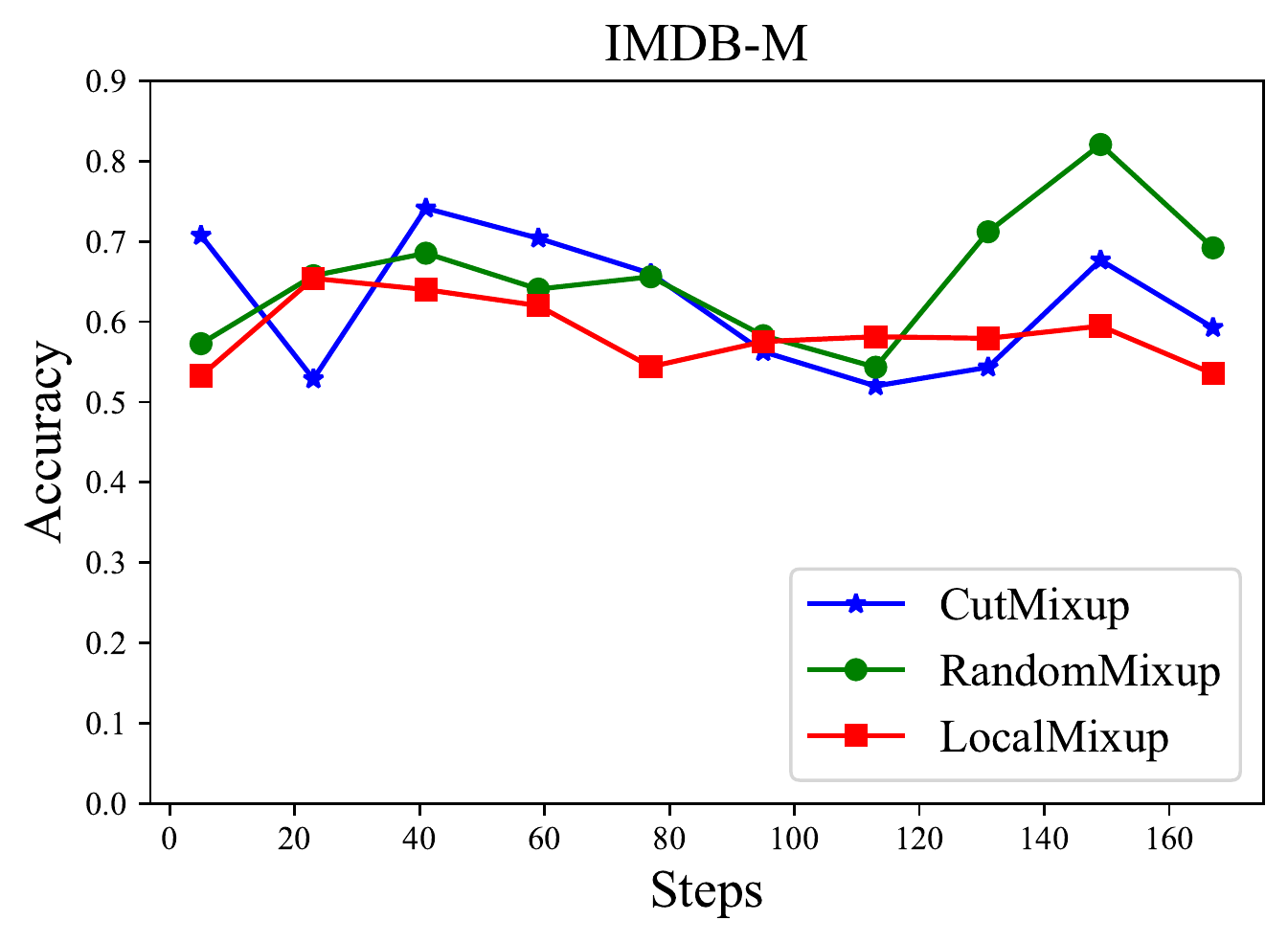}
    \end{minipage}
    }%
    \subfigure[PROTEINS]{
    \begin{minipage}[t]{0.24\linewidth}
    \centering
    \includegraphics[width=1.0\linewidth]{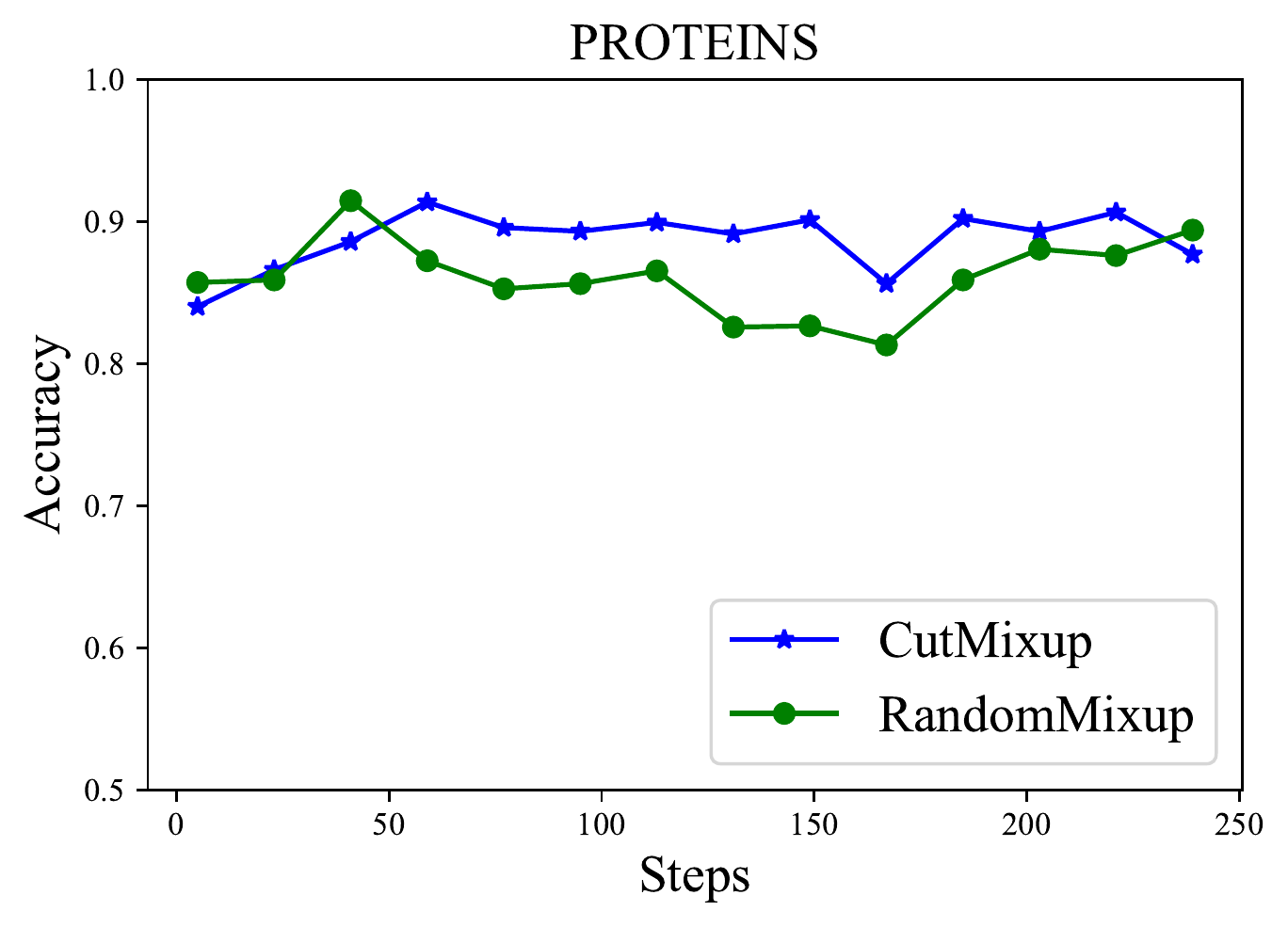}
    \end{minipage}
    }%
    \centering
    \caption{The analytic experiment compares the performance of different mixup strategies on node-level (e.g., Cora, CiteSeer) and graph-level  (e.g., IMDB-M, PROTEINS) datasets. Here we show the two strategies on the PROTEINS, since LocalMixup exceeded 32GB GPU memory limit.
    }
    \label{abstudy-Mixup-Strategy}
\end{figure*}
\begin{figure*}[htbp]
    \centering
    \subfigure[Cora]{
    \begin{minipage}[t]{0.24\linewidth}
    \centering
    \includegraphics[width=1.0\linewidth]{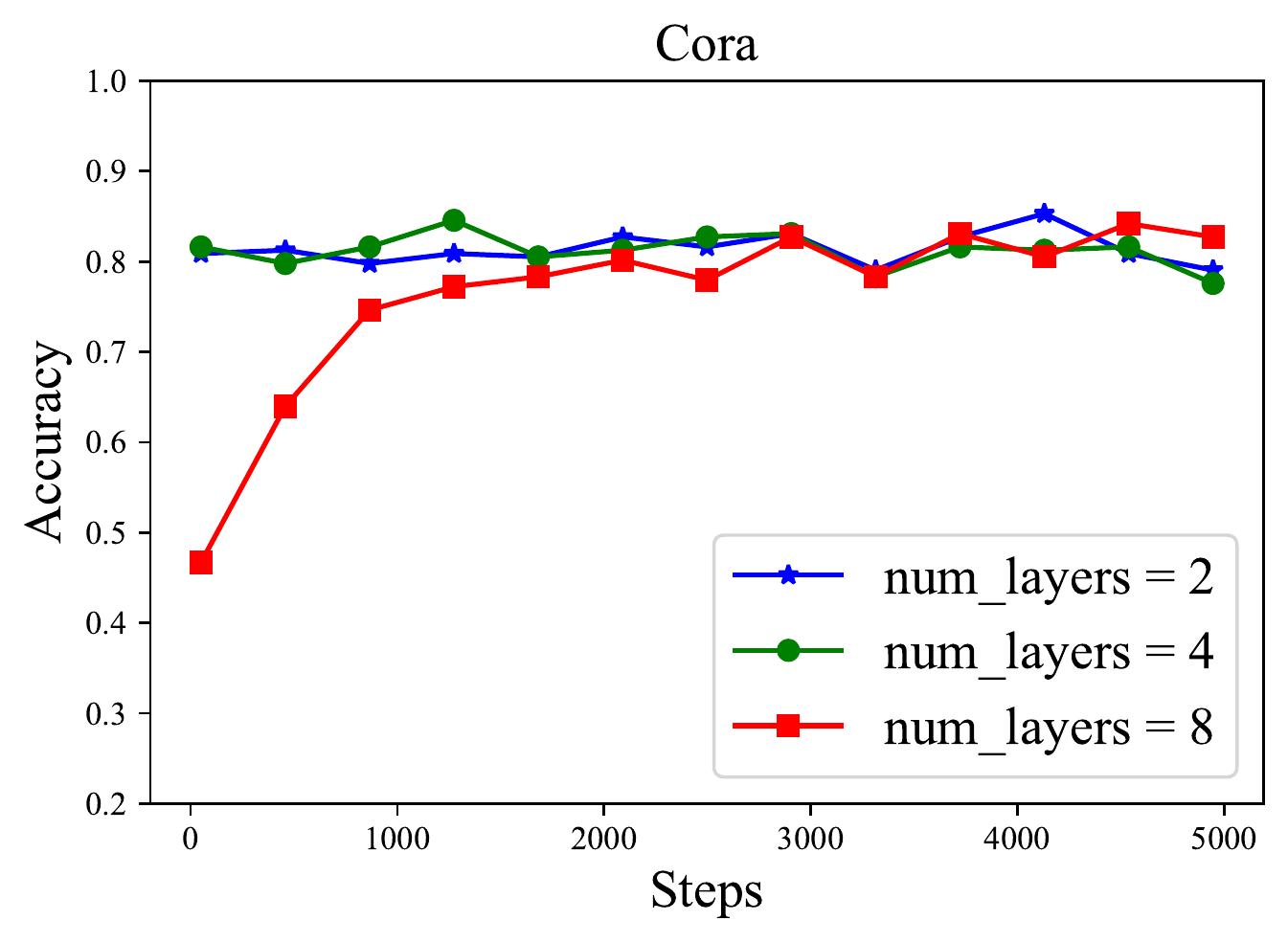}
    \end{minipage}%
    }%
    \subfigure[Citeseer]{
    \begin{minipage}[t]{0.24\linewidth}
    \centering
    \includegraphics[width=1.0\linewidth]{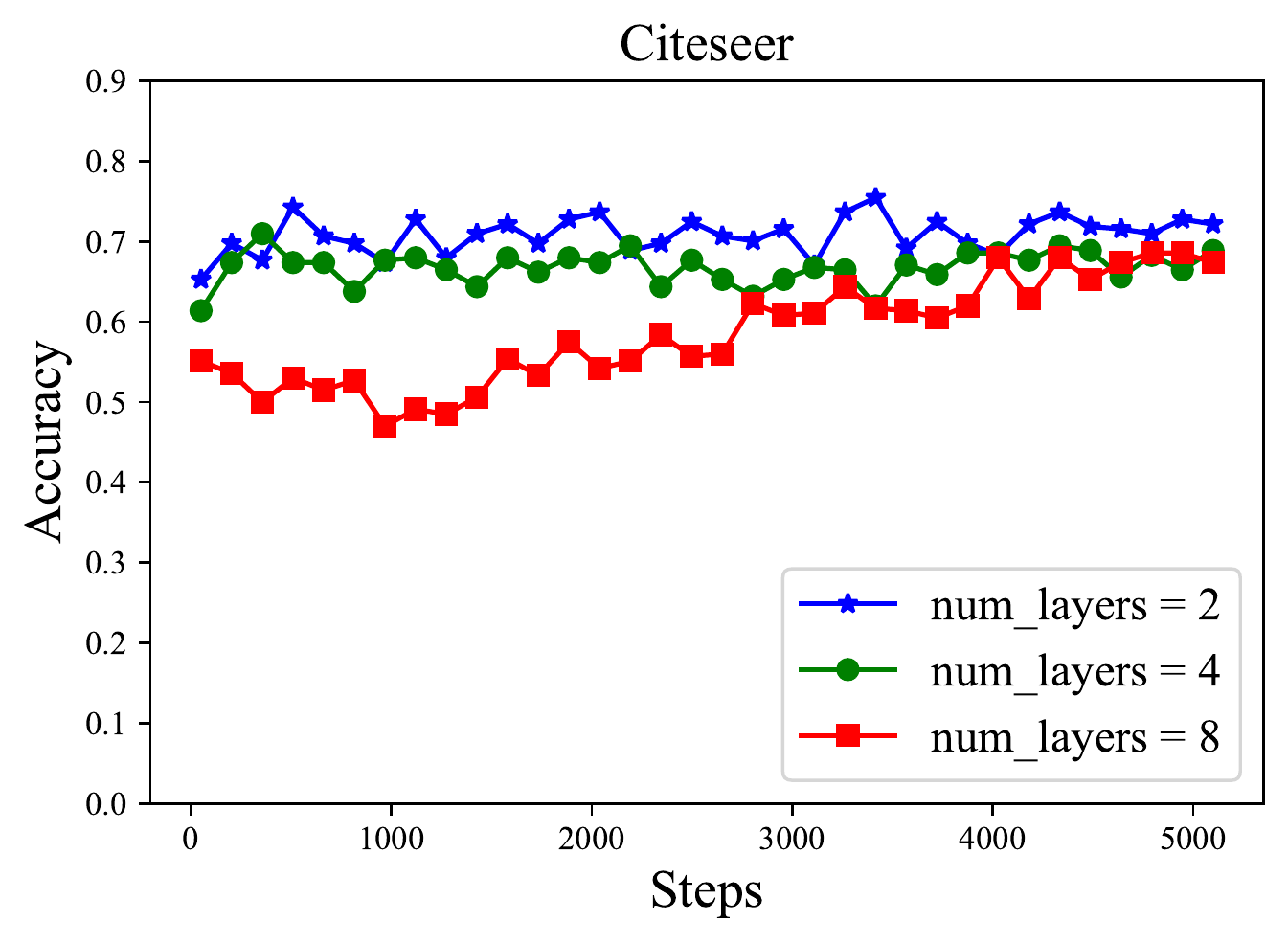}
    \end{minipage}%
    }%
    \subfigure[IMDB-M]{
    \begin{minipage}[t]{0.24\linewidth}
    \centering
    \includegraphics[width=1.0\linewidth]{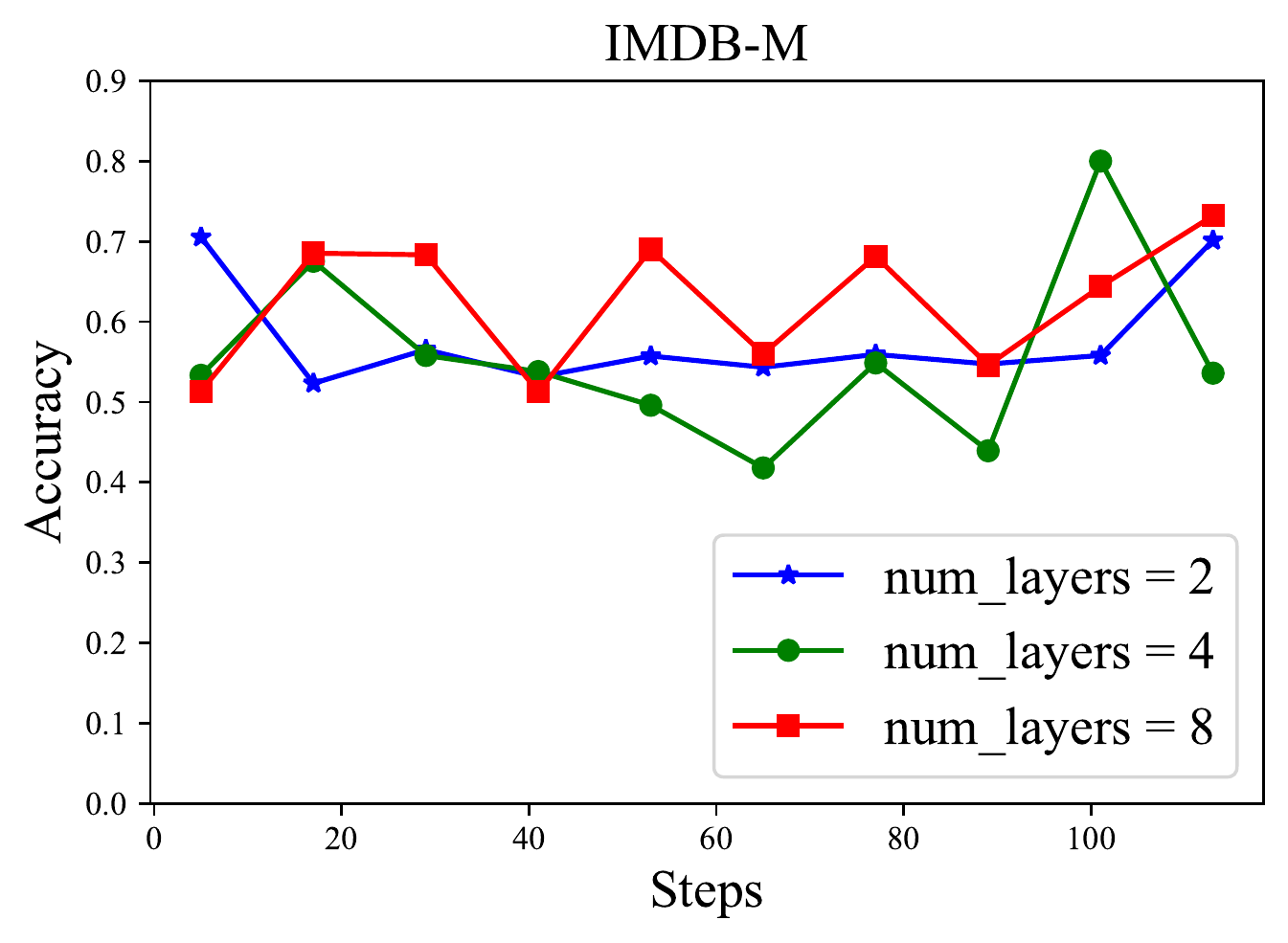}
    \end{minipage}
    }%
    \subfigure[PROTEINS]{
    \begin{minipage}[t]{0.24\linewidth}
    \centering
    \includegraphics[width=1.0\linewidth]{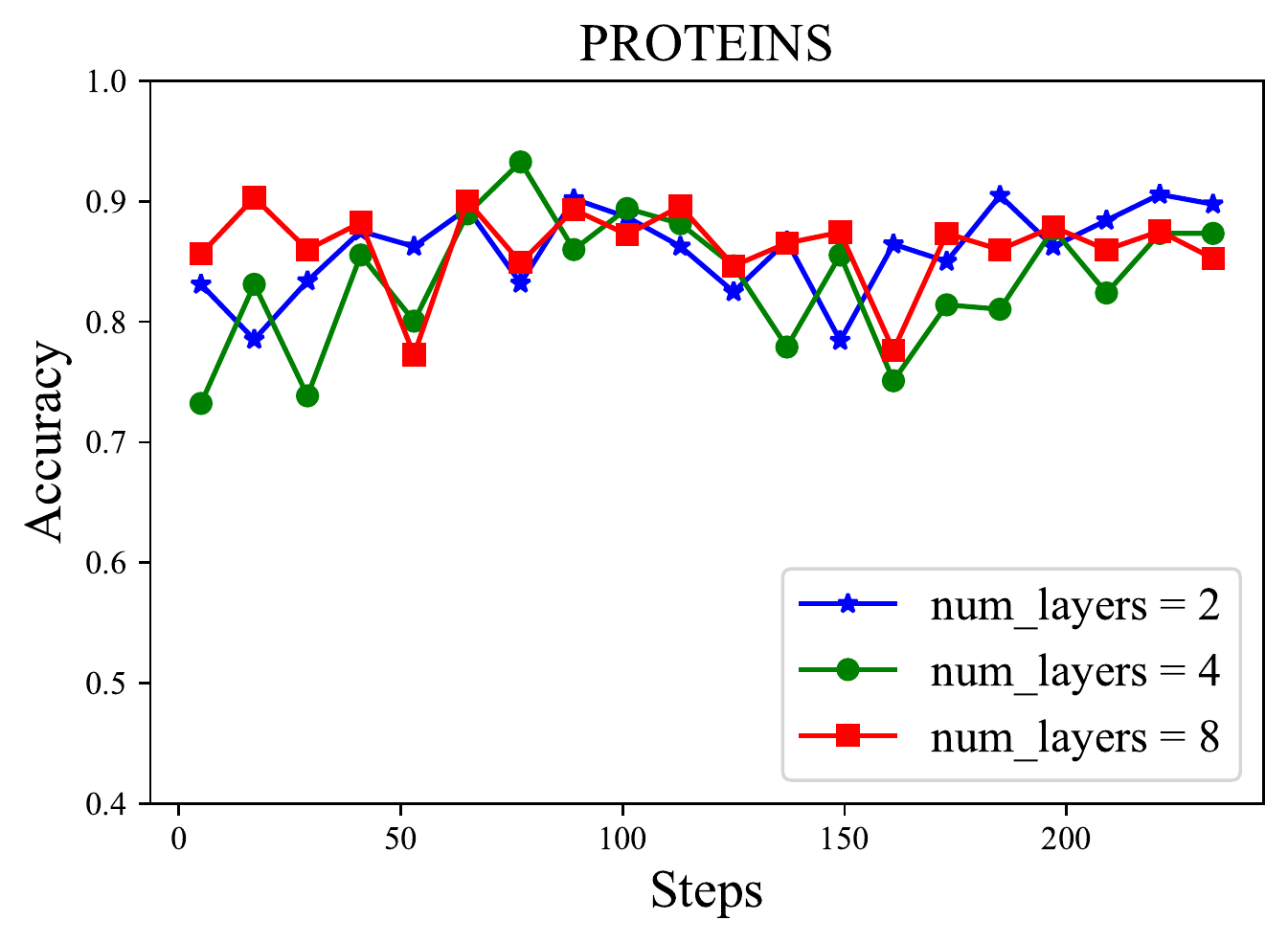}
    \end{minipage}
    }%
    \centering
    \caption{ The analytic experiment compares the performance of the GCN encoder with 2/4/8 layers on node classification datasets (e.g., Cora, CiteSeer) and graph classification datasets (e.g., IMDB-M, PROTEINS). }
    \label{abstudy-n-layers}
\end{figure*}

\subsubsection{Alignment and Uniformity Analysis (Q2)}\label{sec:q2}
To measure the quality of representations obtained from contrastive learning, Wang et al.~\cite{wang2020hypersphere} define two metrics widely used in recent studies~\cite{JunXia2022SimGRACEAS}. One is the alignment metric following the idea that positive samples should stay close in the embedding space:
\begin{align}
\small
    \mathcal{L}_\text{align}(f;\alpha)\triangleq\underset{(\bm{x}_1,\bm{x}_2)\backsim p_{pos}}{\mathbb{E}}[\Vert f(\bm{x}_1)-f(\bm{x}_2)\Vert^\alpha_2]
\end{align}
where $p_{pos}$ is the distribution of positive pairs, and $\alpha>0$. The other is the uniformity metric which is defined as the logarithm of the average pairwise Gaussian potential:
\begin{align}
\small
    \mathcal{L}_\text{uniform}(f;t)\triangleq\log\underset{(\bm{x}_1,\bm{x}_2)\backsim p_{data}}{\mathbb{E}}[e^{-t\Vert f(\bm{x}_1)-f(\bm{x}_2)\Vert^{2}_2}]
\end{align}
where $p_{data}$ is the distribution of data, and $t>0$.
Over the training 1000 epochs, we checked GRACE, MVGRL, COSTA and ID-MixGCL every 2 epochs and visualized alignment $\mathcal{L}_{\text{align}}$ and uniformity $\mathcal{L}_{\text{uniform}}$. 
As depicted in Figure~\ref{align-uniform-visual}, it can be seen that all four approaches enhance the alignment and uniformity of the representation. However, MVGRL achieves a lesser improvement in alignment compared to the others. In contrast, ID-MixGCL attains superior uniformity while simultaneously enhancing alignment, compared to GRACE. Furthermore, while COSTA may attain superior uniformity, its alignment is not as good as that of our method. Thus, our approach is superior overall.
This implies that through the use of identity mixup, ID-MixGCL captures fine-grained representations from unlabeled graphs and learns more precise and smooth decision boundaries on latent features.

\subsubsection{Performance w.r.t Mixup Strategy (Q3)}\label{sec:q3}
We investigated the impact of applying three different mixup strategies on the performance of the model by freezing the encoder output, and the experimental findings are presented in the Figure~\ref{abstudy-Mixup-Strategy}. Our conclusions are as follows: (1) The use of all three mixup strategies was found to enhance the performance of the model on two node-level datasets, such as Cora and Citeseer. Furthermore, the two mixup strategies demonstrated a significantly higher level of effectiveness compared to the baseline on graph-level datasets like IMDB-M and PROTEINS. This indicates the effectiveness of the proposed method. (2) On the node-level datasets, the LocalMixup strategy showed the best performance on the Cora dataset, while the RandomMixup strategy demonstrated the best performance on the CiteSeer dataset. On the graph-level datasets, the RandomMixup strategy performed better on the IMDB-M dataset, while the Cutmixup strategy performed better on the PROTEINS dataset. This partly demonstrates the importance of selecting an appropriate mixup strategy.

\subsubsection{Better Representations alleviate Over-Smoothing (Q4)} \label{sec:q4}
In our study, we utilized a Graph Convolutional Network (GCN) as the backbone for our Graph Neural Network (GNN) model and evaluated its performance using varying numbers of GCN layers - 2, 4, and 8 - across four different datasets. The results of these experiments are illustrated in Figure \ref{abstudy-n-layers}. 
Previous research has shown that the effectiveness of GNN models tends to deteriorate as the number of layers increases~\cite{KaiGuo2021OrthogonalGN}. However, our model was able to partially overcome this performance degradation, achieving comparable results between models with 2 and 8 layers. This suggests that our model may be able to effectively handle a larger number of GCN layers without experiencing a significant decline in performance. We attribute this to the use of an identity mixup technique, which improves the quality of node and graph representations. This claim is also supported in the visual analysis of the previous  subsection.

\begin{figure*}[htbp]
    \centering
    \subfigure[GraphCL+IMDB-B]{
        \begin{minipage}[t]{0.32\linewidth}
        \centering
        \includegraphics[width=0.9\linewidth]{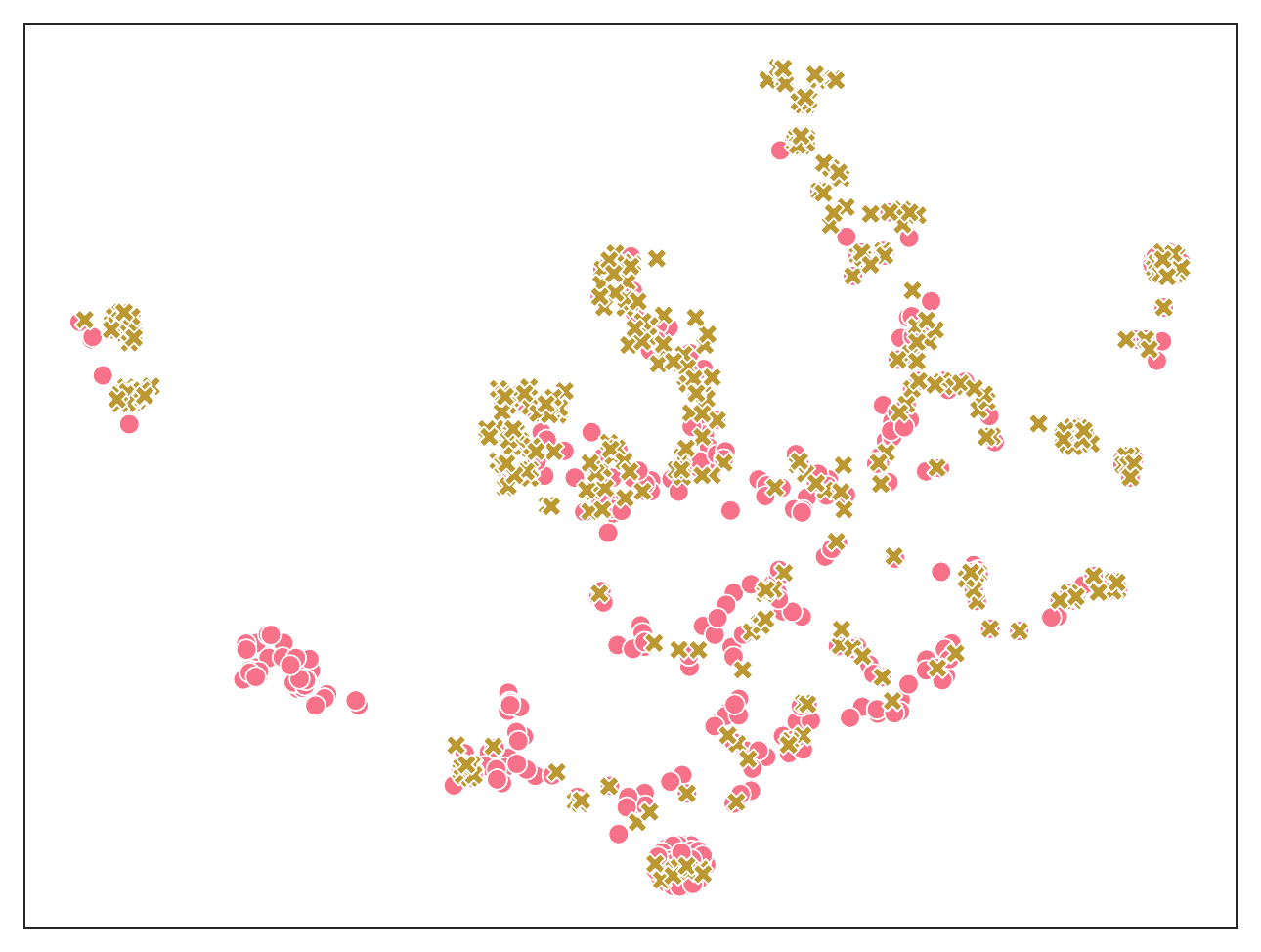}
    \end{minipage}%
    }%
    \subfigure[GraphCL+IMDB-M]{
        \begin{minipage}[t]{0.32\linewidth}
        \centering
        \includegraphics[width=0.9\linewidth]{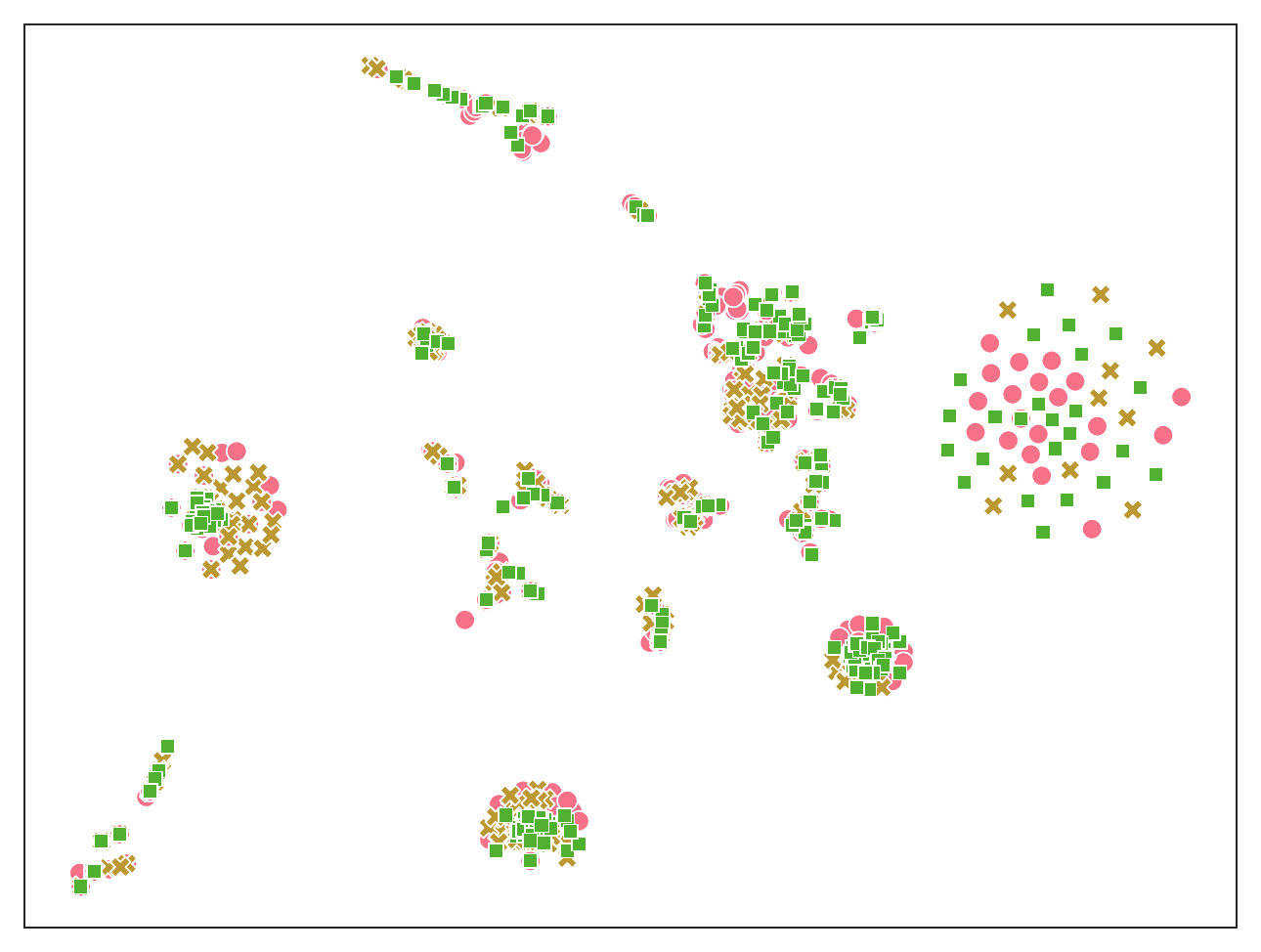}
    \end{minipage}%
    }%
    \subfigure[GraphCL+PROTEINS]{
    \begin{minipage}[t]{0.32\linewidth}
        \centering
        \includegraphics[width=0.9\linewidth]{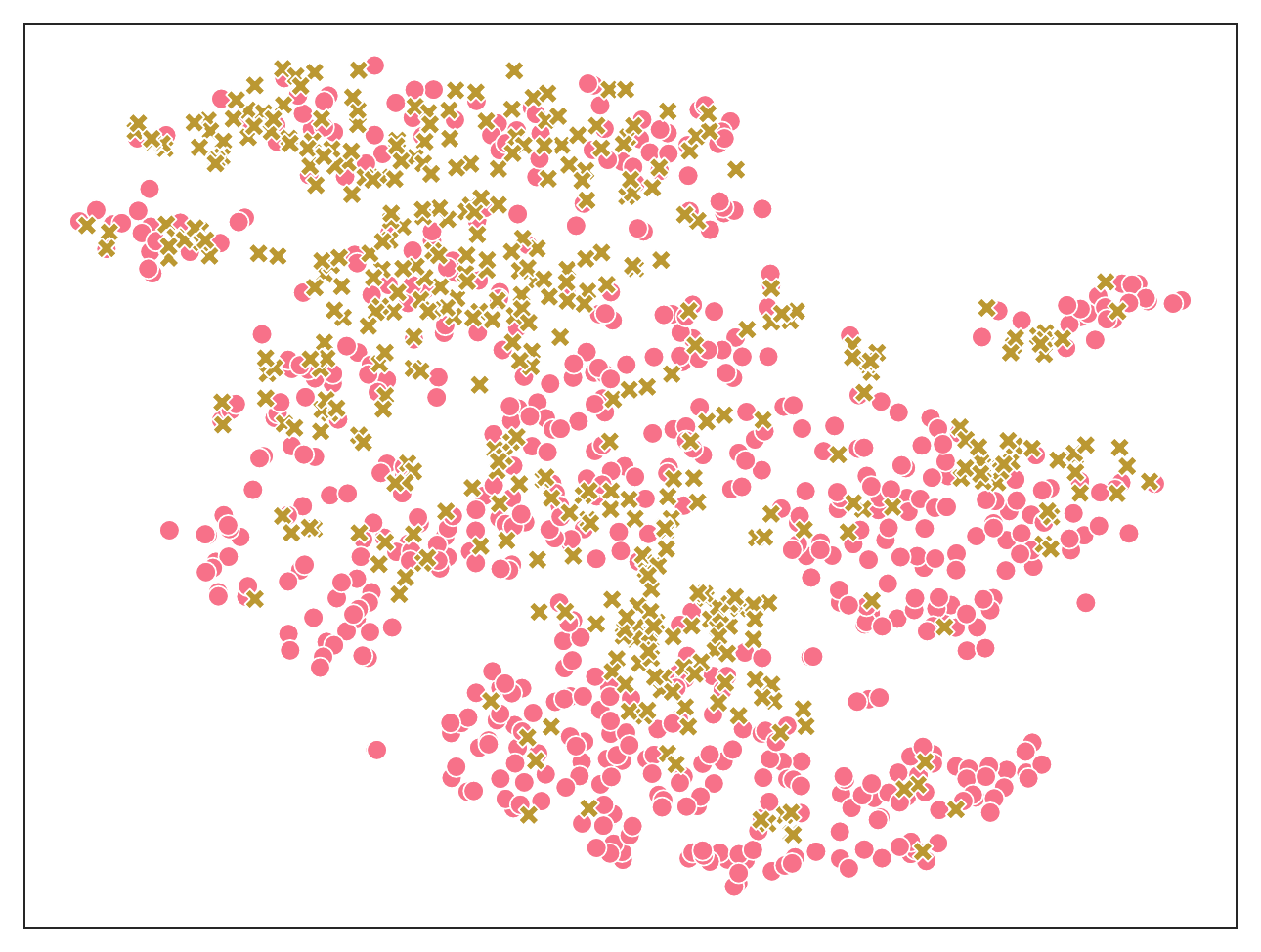}
    \end{minipage}
    }%
    \quad
    \subfigure[ID-MixGCL+IMDB-B]{
    \begin{minipage}[t]{0.32\linewidth}
        \centering
        \includegraphics[width=0.9\linewidth]{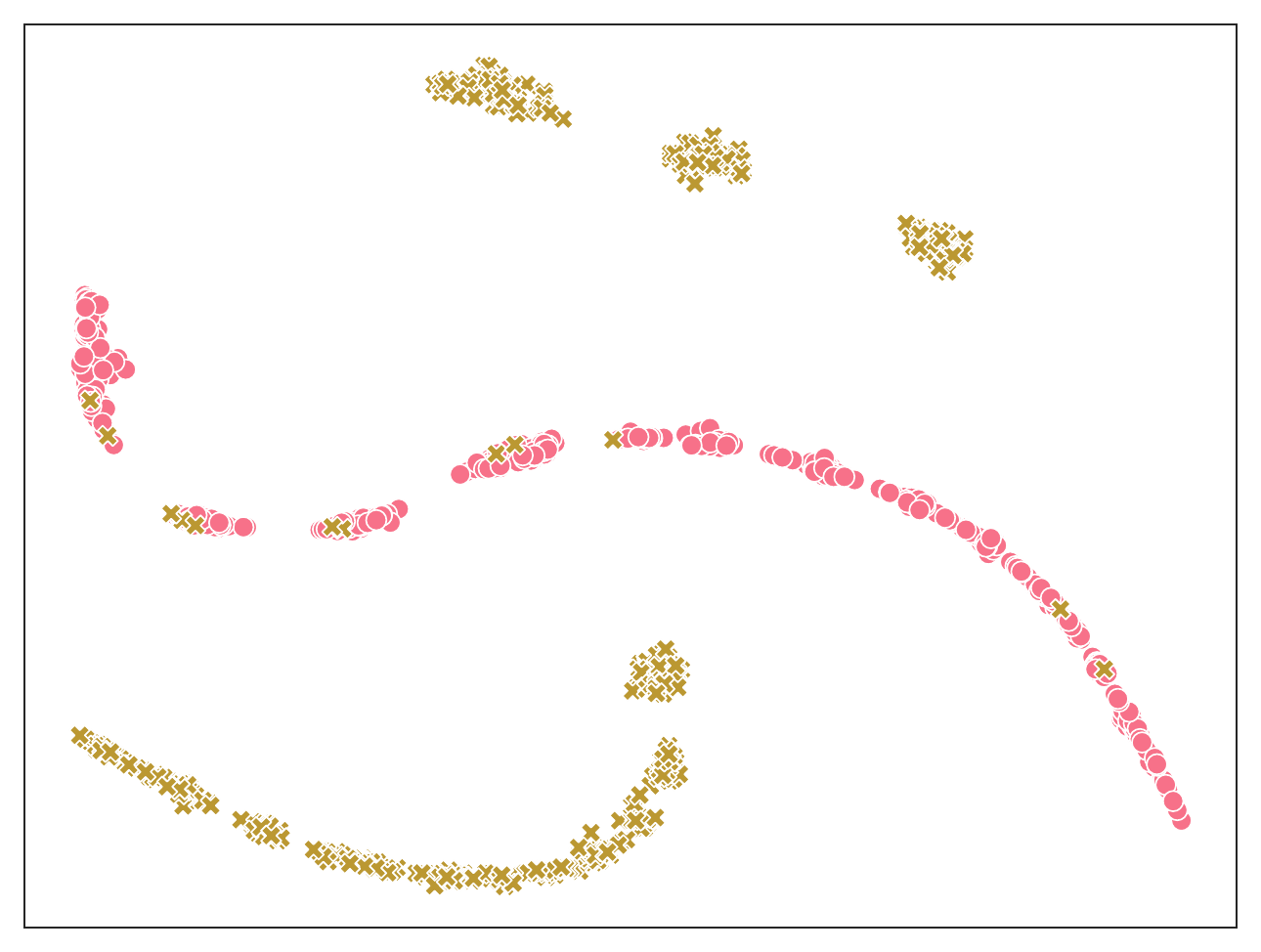}
    \end{minipage}
    }%
    \subfigure[ID-MixGCL+IMDB-M]{
    \begin{minipage}[t]{0.32\linewidth}
        \centering
        \includegraphics[width=0.9\linewidth]{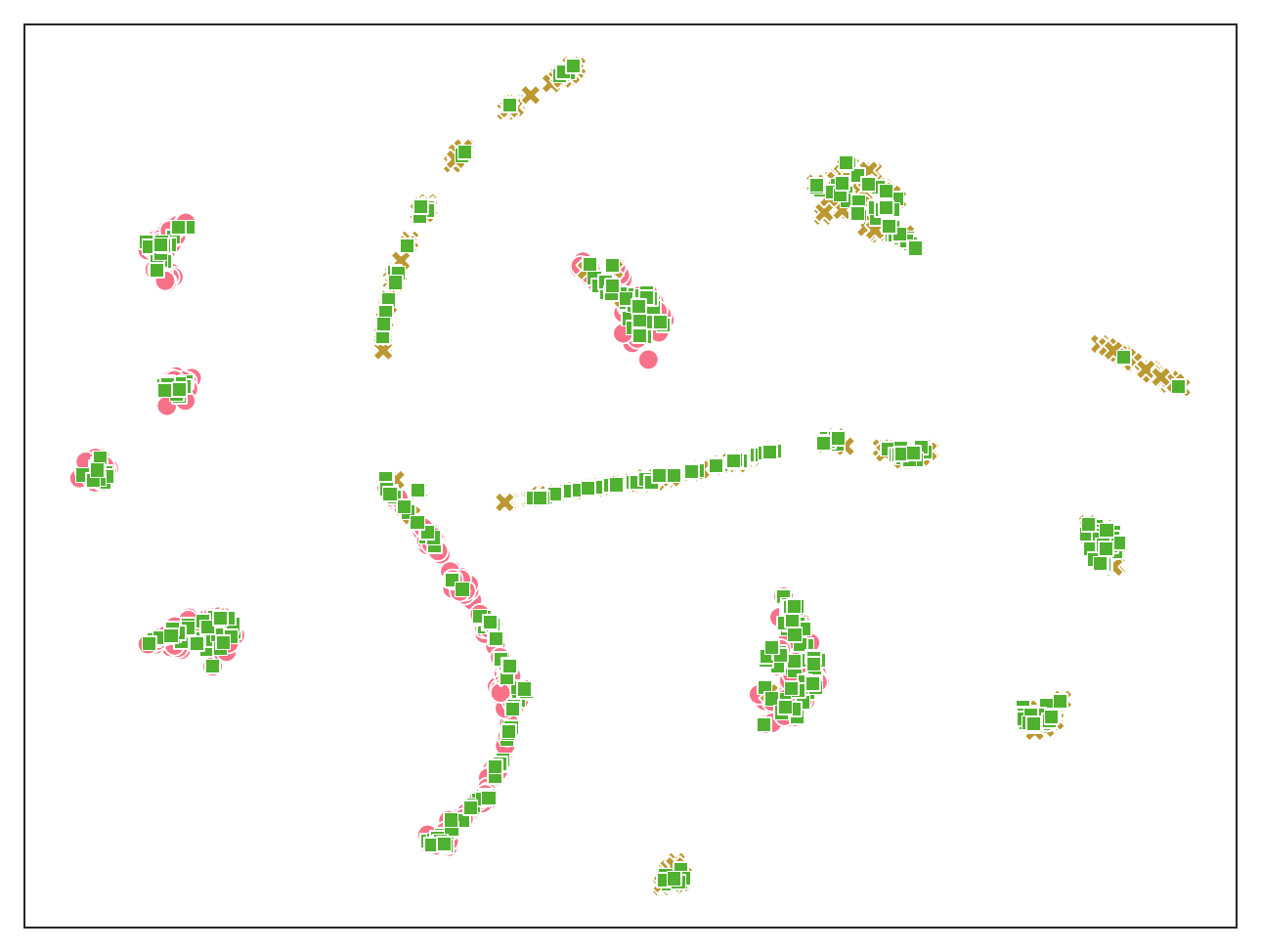}
    \end{minipage}
    }%
    \subfigure[ID-MixGCL+PROTEINS]{
    \begin{minipage}[t]{0.32\linewidth}
        \centering
        \includegraphics[width=0.9\linewidth]{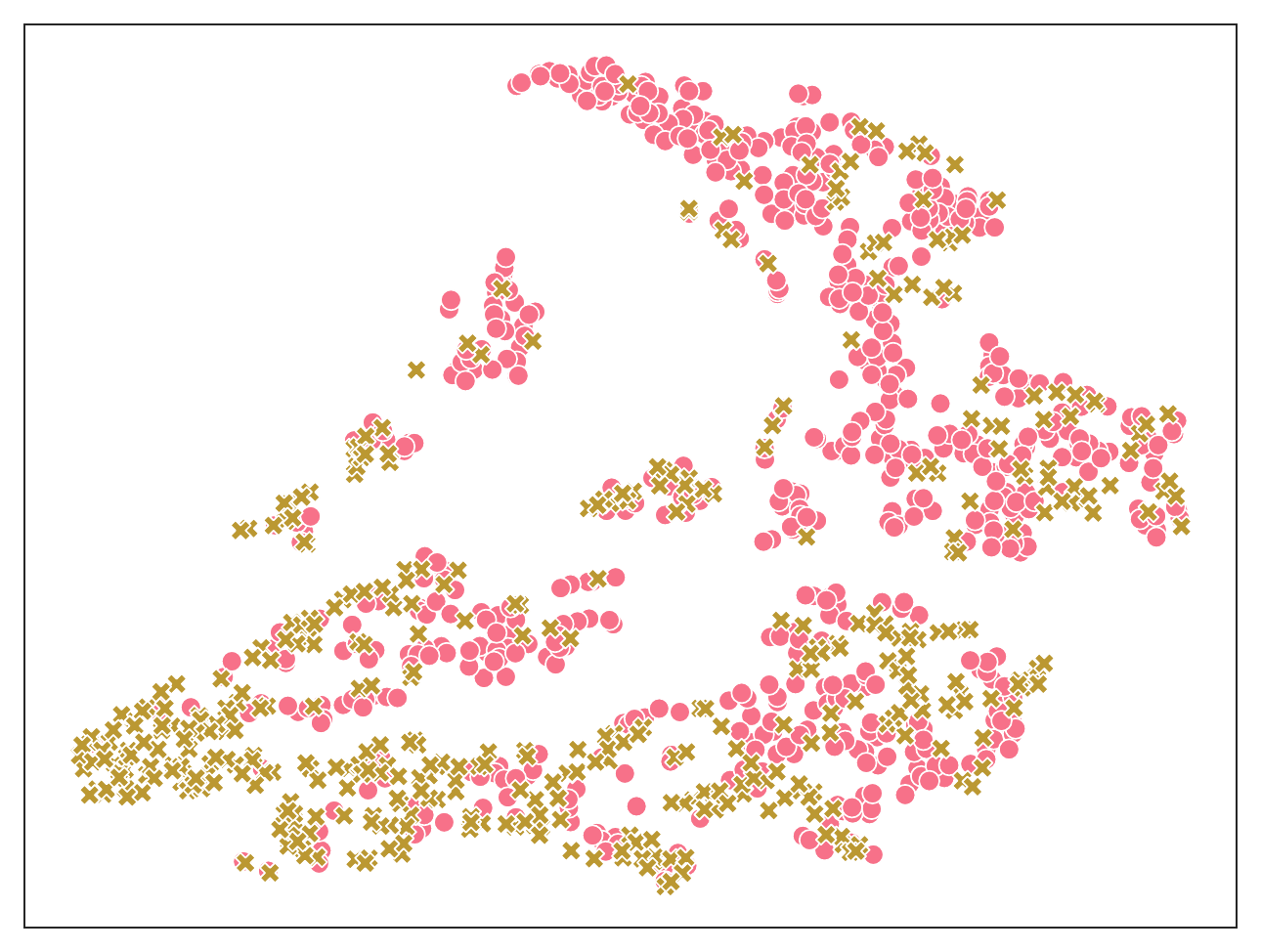}
    \end{minipage}
    }%
    \centering
    \caption{ T-SNE visualization of the learned representation on three graph datasets, where each color represents a class.}
    \label{fig:tsne}
\end{figure*}

\subsubsection{Performance w.r.t Views (Q5)}\label{sec:q5}
In order to evaluate the impact of utilizing multiple view generators, an experimental study was conducted in which the utilization of two distinct view generators was replaced by a single view generator. The results of this experiment are presented in Table \ref{abstudy-view}, which indicate that the use of a multi-view generator is more efficacious than the use of a single-view generator on four datasets, two of which are graph-level datasets and the other two are node-level datasets. 
The observed improvement in performance is attributed to the capability of the multi-view generator to provide a more diverse representation of the same node,  which helps the encoder learn more subtle differences.
\begin{table}[tb]
\footnotesize
    \centering
    \caption{The performance comparison of single-view \textit{vs.} multi-view on node/graph classification datasets. }
    \resizebox{\linewidth}{!}{\begin{tabular}{lcccc}
    \toprule
        & Cora & Citeseer& IMDB-M & PROTEINS \\ \cmidrule(r){2-2} \cmidrule(r){3-3} \cmidrule(r){4-4} \cmidrule(r){5-5} 
        & Accuracy(\%)  & Accuracy(\%)  & Accuracy(\%) & Accuracy(\%) \\ 
        \midrule
        Multi-View & 87.1 ± 0.1  & 75.4 ± 0.2  & 72.25 ± 3.30 & 89.95 ± 2.20 \\
        Single-View & 84.9 ± 0.2 & 74.7 ± 0.1  & 70.53 ± 1.10 & 88.24 ± 1.40 \\ 
        \bottomrule
    \end{tabular}}
    \label{abstudy-view}
\end{table}

\subsubsection{Representation Visualization Compare to GraphCL (Q6)}\label{sec:q6}
To visually show the superior quality of the graph representations learned by ID-MixGCL, we employ t-SNE for the visualization and comparison of the graph representations obtained from ID-MixGCL and GraphCL on three datasets (IMDB-B, IMDB-M, and PROTEINS). Under the identical hyperparameter configuration (50 perplexity with 1000 iterations), we project these high-dimensional representations into a two-dimensional latent space and present the visualization results in Figure \ref{fig:tsne}. As observed from the figure: (1) On the IMDB-B dataset, while GraphCL is capable of identifying some categories, the boundaries between different categories are not as distinct as those obtained by ID-MixGCL. The proposed model achieves superior intra-cluster compactness and inter-cluster separability. (2) On the IMDB-M dataset, while the performance of both GraphCL and ID-MixGCL is subpar, ID-MixGCL demonstrates superior boundary clarity between categories compared to GraphCL. Note that we show that the graph embeddings are not fine-tuned on downstream tasks. After supervised training on a small amount of labeled data, ID-MixGCL outperforms the graph classification accuracy on IMDB-M by more than 20 percentage points.  (3) On the PROTEINS dataset, it is evident that the representations acquired by GraphCL are in a state of relative disorder, with inter-category boundaries not as well-defined as those obtained by ID-MixGCL. Overall, the graph embeddings obtained by our method produce manifolds that are more compact within clusters and more separable between clusters by t-SNE visualization than GraphCL, which we believe is the reason for the huge gains on these datasets.

\section{Conclusion}
In this paper, we propose for the first time that in graph contrastive learning, changes should be made in both the input space and label space, rather than just augmenting the features, as this can lead to the problem of false positives and introduce bias in the model training. 
We propose ID-MixGCL, which generates soft similarity samples by interpolating both node representations and identity labels, helping the encoder learn better representations. We validate its effectiveness on node and graph classification tasks across 14 datasets, and analysis experiments show that our method helps the encoder learn representations with better uniformity and alignment. 
T-SNE visualization comparisons with other methods demonstrate that our method can produce better manifold distributions, which is the fundamental reason for the gains of our method. 
Although simple, we hope that ID-MixGCL will become a widely accepted baseline method for GCL in the future.


\section*{ACKNOWLEDGMENTS}
We would like to thank the anonymous reviewers for their comments. 
This work was supported by the Strategic Priority Research Program of Chinese Academy of Sciences under Grant No.XDC02040400, the Youth Innovation Promotion Association of CAS under Grant No.2021153.

\bibliographystyle{IEEEtran}
\bibliography{main.bib}


\end{document}